%% file: main.tex
\documentclass[letterpaper]{article} 
\usepackage{aaai2026}  
\usepackage{times}  
\usepackage{helvet}  
\usepackage{courier}  
\usepackage[hyphens]{url}  
\usepackage{graphicx} 
\usepackage{natbib}  
\usepackage{caption} 
\usepackage{algorithm}
\usepackage{algorithmic}
\usepackage{amsmath}
\usepackage{multirow}
\usepackage{array}
\usepackage{colortbl}
\usepackage[dvipsnames,table]{xcolor}
\usepackage{fontawesome5}
\usepackage{pifont}
\usepackage{amssymb}
\usepackage{enumitem}
\usepackage{subcaption} 

\usepackage{newfloat}
\usepackage{listings}
\DeclareCaptionStyle{ruled}{labelfont=normalfont,labelsep=colon,strut=off} 
\floatstyle{ruled}
\newfloat{listing}{tb}{lst}{}
\floatname{listing}{Listing}

\title{Exo2Ego: Exocentric Knowledge Guided MLLM for Egocentric \\ Video Understanding}
\author{
    Haoyu Zhang\textsuperscript{\rm 1,2}, Qiaohui Chu\textsuperscript{\rm 1,2},
    Meng Liu\textsuperscript{\rm 3,4$^*$},
    Haoxiang Shi\textsuperscript{\rm 1,2},
    Yaowei Wang\textsuperscript{\rm 1,2$^*$},
    Liqiang Nie\textsuperscript{\rm 1\thanks{Corresponding authors.}}
}
\affiliations{
    \textsuperscript{\rm 1}Harbin Institute of Technology (Shenzhen) \\
    \textsuperscript{\rm 2}Pengcheng Laboratory\\
    \textsuperscript{\rm 3}Shandong Jianzhu University \\ \textsuperscript{\rm 4}Zhongguancun Academy\\

    \{zhang.hy.2019, qiaohuichu8599,     mengliu.sdu, shihaoxiang1999, nieliqiang\}@gmail.com\\
    wangyw@pcl.ac.cn
}

\usepackage{bibentry}

\begin{document}

\maketitle

\input{sec/0_abstract}    
\input{sec/1_intro}
\input{sec/2_relatedwork}

\input{sec/3_data}
\input{sec/4_method}

\input{sec/5_experiment}

\input{sec/6_conclusion}

\input{sec/X_suppl}


{
\small
\bibliography{main}
}

\end{document}

%% file: sec/0_abstract.tex
\begin{abstract}
AI personal assistants, deployed through robots or wearables, require embodied understanding to collaborate effectively with humans. However, current Multimodal Large Language Models (MLLMs) primarily focus on third-person (exocentric) vision, overlooking the unique challenges of first-person (egocentric) videos. Additionally, high acquisition costs limit data size, impairing MLLM performance.
To address these challenges, we propose learning the mapping between exocentric and egocentric domains, leveraging the extensive exocentric knowledge within existing MLLMs to enhance egocentric video understanding. To this end, we introduce Ego-ExoClip, a pre-training dataset comprising 1.1M synchronized ego-exo clip-text pairs derived from Ego-Exo4D, together with the instruction-tuning dataset EgoIT, which is collected from multiple sources to enhance the model's instruction-following capabilities. Building upon the datasets, we propose a migration strategy and further design a progressive mapping learning pipeline with three stages: Demonstrator Self-Preparation, Demonstrator-Learner Guidance, and Learner Self-Practice. 
Extensive experiments across diverse egocentric tasks reveal that existing MLLMs perform inadequately in egocentric video understanding, while our model significantly outperforms these leading models. 
\end{abstract}

\begin{links}
    \link{Code}{https://reurl.cc/Ebpyrm}
\end{links}

%% file: sec/1_intro.tex
\section{Introduction}

\begin{figure}[t]
  \centering
   \includegraphics[width=\linewidth]{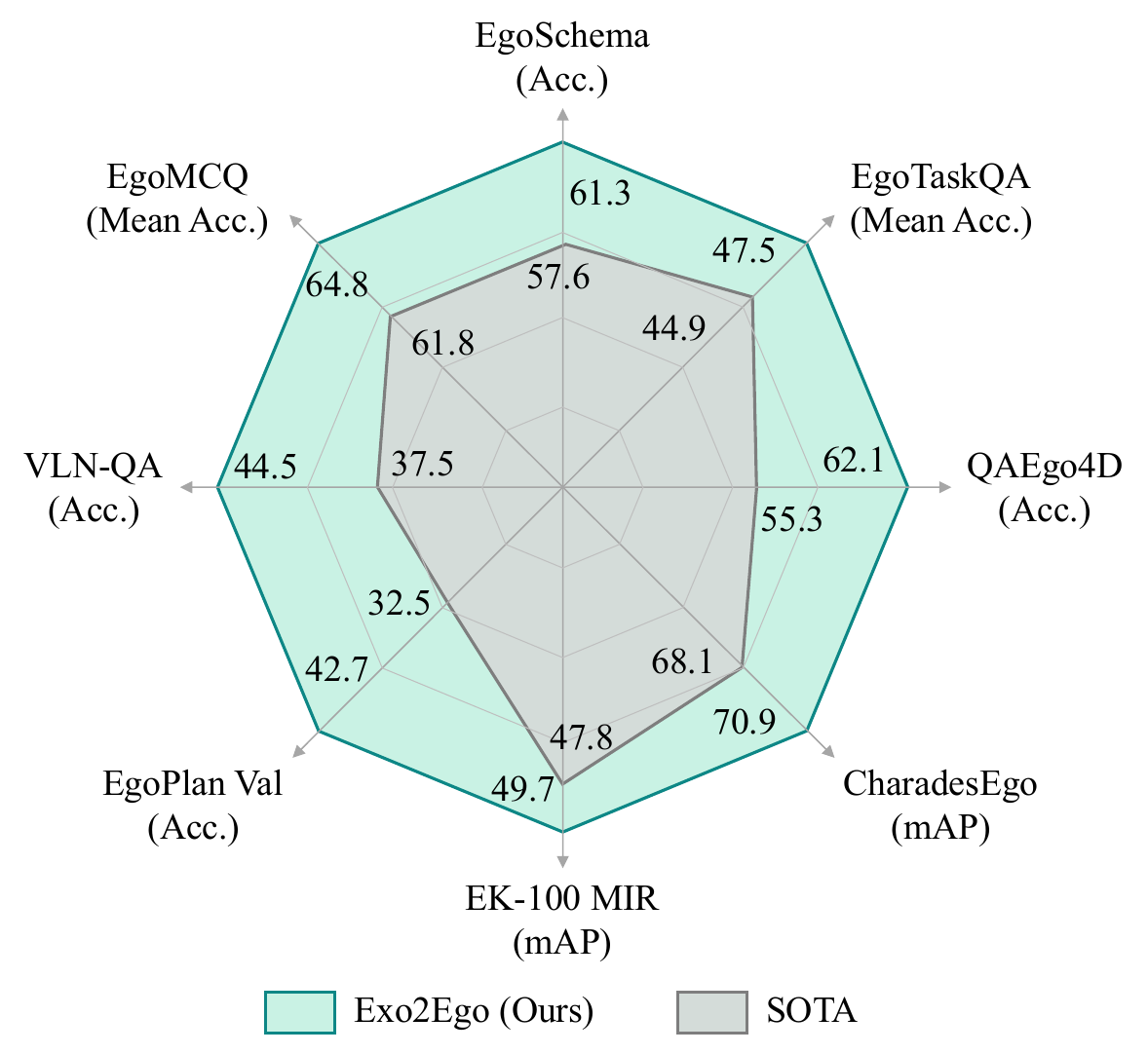}
   \caption{Our Exo2Ego model achieves optimal results across various egocentric video understanding tasks, with the detailed results presented in Table~\ref{tab:vqa}.}
   \label{fig:intro}
\end{figure}

\textit{An outsider can see things more clearly or objectively than those involved.\\ \rightline{-- The Old Book of Tang}\\}

\noindent Embodied cognition theory posits that cognitive processes are fundamentally shaped by interactions with the physical environment~\cite{10.3389/fpsyg.2015.00875}. In this context, understanding egocentric videos that capture human active perception and fine-grained action from a first-person perspective provides a vital pathway for artificial intelligence to understand human experiences. The challenge of analyzing egocentric videos arises from the dynamic interplay between the movements of the camera-wearer (i.e., self) and the surrounding environment (i.e., other), which distinguishes them from exocentric videos observed from a third-person perspective, where this direct experience is absent. 
The potential applications of this technology span a wide range, including visual aids, smart glasses, and immersive experiences in virtual and augmented reality. 
 

In response to the increasing demand for effective egocentric video analysis, several research tasks have emerged, such as egocentric action recognition~\cite{shiota2024egocentric,pmlr-v235-zhang24aj}, episodic memory~\cite{feng2024objectnlq,wang2022siamese,Feng_2025_CVPR}, human-object interaction~\cite{jiang2023full}, and action anticipation~\cite{chu2025intention,chu2025technical}. \textit{However, the diverse architectures employed across these tasks lead to significant fragmentation in the field.} With the success of existing Multimodal Large Language Models (MLLMs)~\cite{wang2025time, zhang2025spatial}, there is an emerging trend toward developing a unified egocentric MLLM architecture capable of addressing multiple tasks within the egocentric domain.
A significant obstacle in this direction is the limited availability of egocentric video data, which is crucial for effectively training large-capacity models~\cite{wang2023ego}. 
Consequently, an increasing number of methods~\cite{li2021ego,xue2023learning,luo2025put,dou2024unlocking} are now retrieving suitable exocentric videos from the predefined set to aid in model training for downstream tasks.
While these approaches exhibit empirical advantages in transferring knowledge from exocentric to egocentric representations, \textit{they incur additional retrieval time and suffer from  misalignment, leading to instability in model performance. 
}

\begin{figure}[t]
  \centering
   \includegraphics[width=0.97\linewidth]{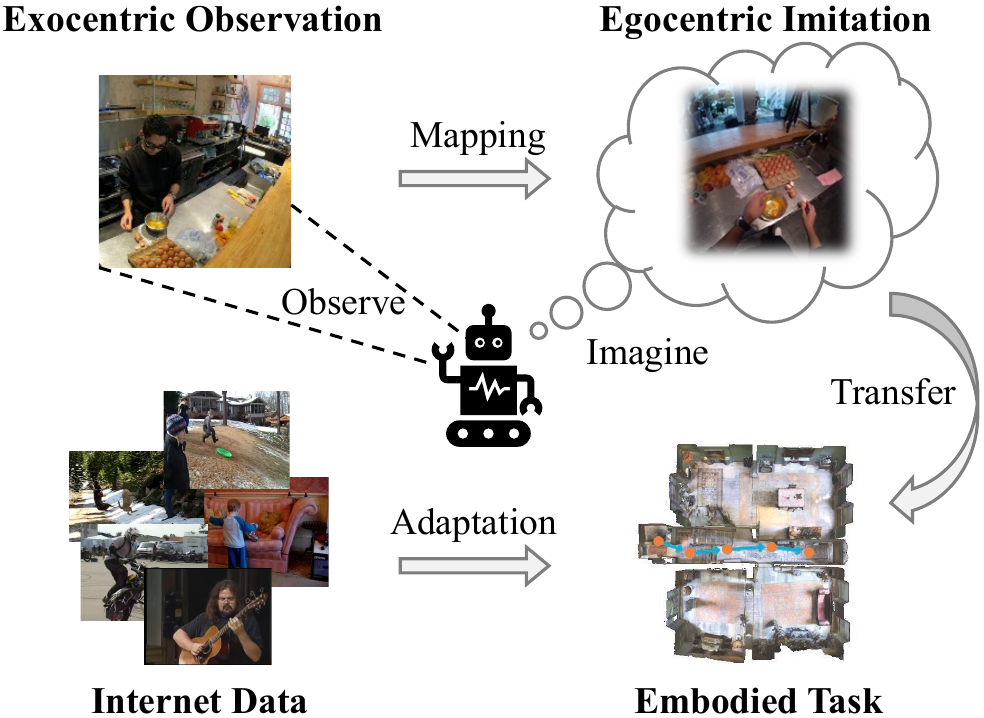}
   \caption{Understanding how humans map observed activities to their own behaviors enhances embodied cognition across various downstream tasks by transferring insights from existing Internet (i.e., exocentric) data.}
   \label{fig:idea}
\end{figure}

Drawing inspiration from cognitive science~\cite{shi2024cognition,zhang2023attribute}, which demonstrates that children learn by observing others’ behavior (i.e., an exocentric view) and mapping these observations onto their own experiences (i.e., an egocentric perspective), we conceptualize the exocentric video observer 
as the ``demonstrator'' and the egocentric video interpreter
as the ``learner''. Our objective is to establish a robust mapping pattern between the two, as illustrated in Figure~\ref{fig:idea}, offering two key advantages: 1) \textit{Weak Dependence.} Leveraging the learned mappings, the model can flexibly tackle downstream tasks without relying on cross-domain data, thereby mitigating additional biases.
2) \textit{Strong Generalization.} By emulating human learning, these mapping techniques significantly enhance the model's generalization capabilities, reducing the need for extensive egocentric training data in MLLMs.

To further this approach, we propose an exocentric-to-egocentric migration strategy with an egocentric self-consistency mechanism grounded in cross-view behavior invariance.
By embedding this strategy within MLLMs, we design a mapping learning pipeline comprising three stages: \textbf{Demonstrator Self-Preparation}, \textbf{Demonstrator-Learner Guidance}, and \textbf{Learner Self-Practice.} By transferring exocentric knowledge, this pipeline boosts egocentric understanding and improves downstream performance. Furthermore, to satisfy the demand for synchronized cross-domain data, we introduce the \textbf{Ego-ExoClip} dataset, an advanced first- and third-person video-text dataset containing 1.1M egocentric-exocentric clip-text pairs. And we also construct an enriched instruction-tuning dataset, \textbf{EgoIT}, sourced from multiple egocentric contexts to improve the instruction-following capabilities.
As shown in Figure~\ref{fig:intro}, abundant results demonstrate that our framework achieves superior results, highlighting its effectiveness in advancing egocentric video understanding.

\noindent\textbf{Contributions}: \textbf{1}) Based on behavior invariance, we propose a knowledge transfer strategy and integrate it into the training paradigm of MLLMs, resulting in a three-stage mapping learning process. \textbf{2}) To support training, we construct a synchronized exo-ego video-text dataset, Ego-ExoClip, as well as an instruction-tuning dataset, EgoIT, collected from multiple sources. \textbf{3}) Extensive experimental results on various benchmarks validate the effectiveness of our proposed datasets and methods, and further promote research in embodied cognition.

%% file: sec/2_relatedwork.tex
\section{Related Work}

\textbf{Egocentric Video Understanding.}
Egocentric videos offer a distinctive perspective for active engagement with the physical world, capturing human interactions from a first-person viewpoint. A variety of datasets have been developed to support research in this field, including EPIC-KITCHENS~\cite{damen2020epic}, which focuses on kitchen activities; Charades-Ego~\cite{sigurdsson2018charades}, which highlights diverse everyday tasks; and Ego4D~\cite{grauman2022ego4d}, which provides a globally diverse collection of egocentric videos. These datasets have driven the growth of multiple research areas, such as video-text pre-training~\cite{lin2022egocentric, pramanick2023egovlpv2}, video question answering~\cite{di2024grounded,zhang2024hcqa,zhang2021multimodal}, human-object interaction~\cite{akiva2023self}, action anticipation~\cite{ragusa2023stillfast,shen2024progress}, episodic memory~\cite{ramakrishnan2023naq,feng2024objectnlq}, and pose perception~\cite{millerdurai2024eventego3d}. 
Despite these advancements, the high cost of data collection remains a major limitation, constraining dataset scale and, in turn, the effectiveness of the models trained on them.
In response, recent work~\cite{wang2023learning,xue2023learning,truong2024cross} has begun exploring the use of unpaired exocentric videos to enhance egocentric video comprehension. However, leveraging unpaired exocentric data introduces inherent biases that can negatively affect model stability and accuracy. To address this issue, we introduce a pairwise egocentric-exocentric dataset with diverse activities and leverage the exocentric knowledge embedded in MLLMs to improve egocentric video understanding.

\noindent \textbf{Multimodal Large Language Model.}
Large Language Models (LLMs)~\cite{touvron2023llama,DBLP:journals/pami/WenSYWGN24}, trained on vast datasets from the Internet, have demonstrated remarkable proficiency across a wide range of language tasks through text generation.  
Building upon the success of LLMs and the promise of multimodal data, there is growing scholarly interest in exploring MLLMs.
Early MLLMs focus on integrating static images with text, with landmark work such as Flamingo~\cite{alayrac2022flamingo} and BLIP-2~\cite{li2023blip}, significantly expanding the capabilities of LLMs for multimodal tasks. Recent open-source MLLMs, including LLaVA~\cite{liu2024visual} and InstructBLIP~\cite{NEURIPS2023_9a6a435e}, further advance visual understanding by introducing visual instruction-tuning data. 
Beyond static images, recent studies have started to explore the incorporation of video data into LLMs, unlocking new potential for video comprehension tasks. 
Innovations such as LLaMA-VID~\cite{li2023llama}, Video-LLaVA~\cite{lin2023video}, and VideoLLaMA2~\cite{cheng2024videollama} aim to improve the alignment of video data with LLMs. Additionally, methods like VideoChatGPT~\cite{maaz2023video} and VideoChat2~\cite{li2024mvbench} utilize ChatGPT to generate video instruction-tuning data, enhancing instruction-following capabilities. 
Despite these advancements, existing MLLMs exhibit limited performance on egocentric video understanding tasks~\cite{wang2023ego}. Therefore, our objective is to unify various downstream tasks and significantly enhance the capabilities of MLLMs in the embodied cognition domain.

%% file: sec/3_data.tex
\section{Dataset Construction}
In this section, we provide a detailed overview of the constructed Ego-ExoClip dataset, which is employed in Stages 1 and 2, along with the instruction-tuning dataset, EgoIT, developed for Stage 3 of the training process.

\begin{figure}[t]
  \centering
   \includegraphics[width=0.93\linewidth]{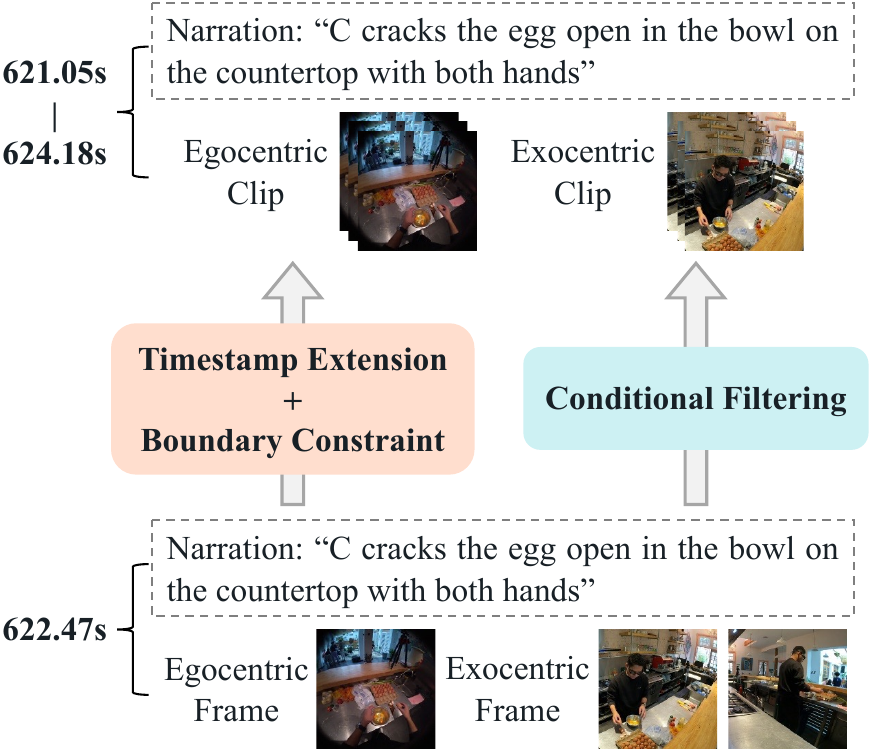}
   \caption{Illustration of the construction of  Ego-ExoClip.}
   \label{fig:ego-exoclip}
\end{figure}

\subsection{Ego-ExoClip}
The Ego-ExoClip dataset is derived from Ego-Exo4D~\cite{grauman2024ego}, which comprises 5,035 grouped videos ranging in length from 1 to 42 minutes.
Each group contains one egocentric video paired with 4–5 simultaneously captured exocentric videos. Most of these videos include dense timestamp-level narrations provided by two annotators, detailing the camera wearer's activities and interactions. For example, as shown in Figure~\ref{fig:ego-exoclip}, the narration ``C cracks the egg open in the bowl on the countertop with both hands'' corresponds to an event occurring at 622.47 seconds, where ``C'' refers to the camera-wearer. 

To ensure data quality and consistency, we filter out videos without narrations, those lacking UID mappings\footnote{UID mapping associate narrations and corresponding videos.}, and those designated for validation or testing in benchmark challenges. 
This process yields a refined dataset of 2,925 video groups, totaling 15,478 videos, all validated to guarantee lossless quality.
For incorporating narration diversity, we retain textual annotations from both narrators, resulting in 623.6 hours of videos and 261.3K narrations.

Given that the narrations in Ego-Exo4D are annotated at single timestamp rather than as time intervals, we extend the timestamp-level narrations to the clip-level. Specifically, narrations are organized as a sequence of sentences $\{\mathcal{S}_0,...,\mathcal{S}_n\}$ with corresponding timestamps $\{t_0,...,t_n\}$, indicating that an event $i$ described by $\mathcal{S}_i$ occurred at time $t_i$, where $n$ denotes the total number.
For a narration $\mathcal{S}_i$ with timestamp $t_i$, we define the corresponding clip $\mathcal{C}_i$ with the following start $t_{i}^{s}$ and end $t_{i}^{e}$ timepoints:

\begin{equation}
\begin{cases}
    t_{i}^{s}=\text{Max} (t_i-\beta_i/2\alpha, t_{i-1}), \\
    t_{i}^{e}=\text{Min} (t_i+\beta_i/2\alpha, t_{i+1}),
\end{cases}
\end{equation}
where $\beta_i$ is a parameter representing the average temporal distance between consecutive narrations, calculated as $\sum_{j=0}^{n-1}(t_{j+1} - t_j)/n$. This value is calculated for each video individually, with $\alpha$ serving as a scale factor derived from the average of all $\beta_i$ across the entire Ego-ExoClip dataset ($\alpha$ = 1.92 seconds). To ensure that a description does not span multiple clips, we utilize the preceding and following timestamps as boundary markers.

\begin{figure*}[t]
  \centering
   \includegraphics[width=0.98\linewidth]{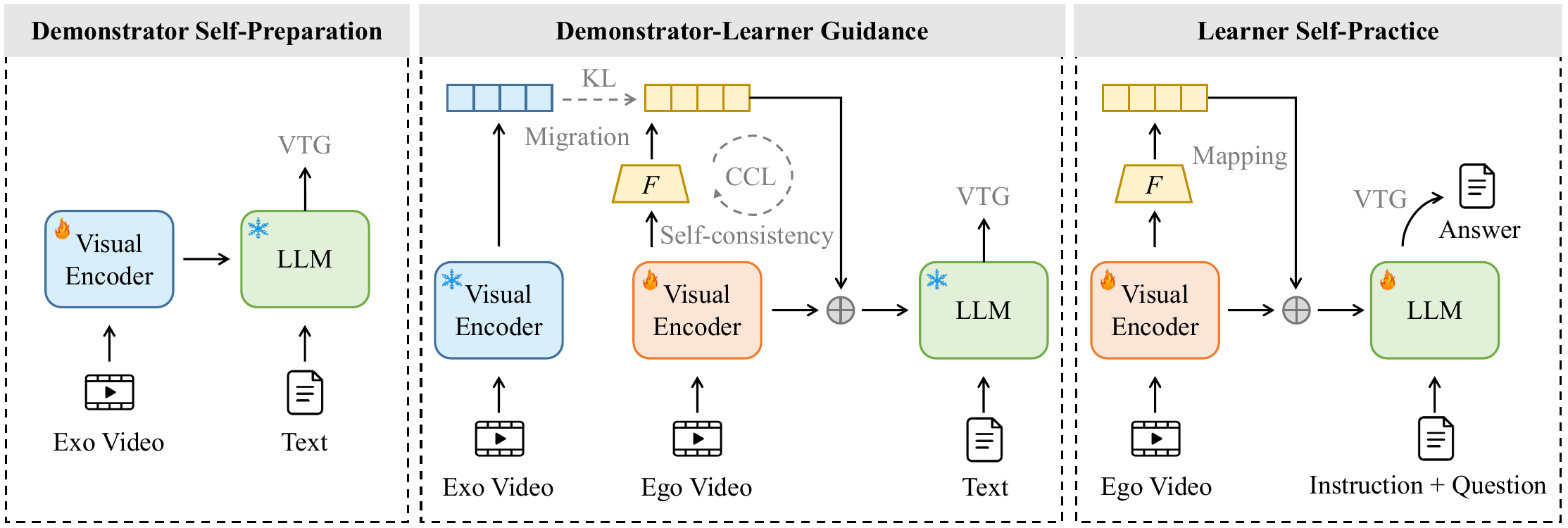}
   \caption{Illustration of our progressive training process, including the stages of Demonstrator Self-Preparation, Demonstrator-Learner Guidance, and Learner Self-Practice. ``VTG", ``KL", and ``CCL" denote vision-grounded text generation, Kullback-Leibler divergence, and cycle consistency loss, respectively. $F$ is a mapping function.}
   \label{fig:method}
\end{figure*}

\subsection{EgoIT}\label{IT_data}
To enhance the instruction-following capabilities of MLLMs for downstream egocentric tasks, we introduce the EgoIT dataset, which comprises approximately 600K samples sourced from multiple distinct domains. All data samples are standardized into a uniform format. Each sample consists of four main components: the video path, task instruction, question, and answer. To ensure diversity in instructions, we integrate the ``dataset description", ``task description", and ``instruction example" information into a prompt template, using GPT-4o\footnote{\url{https://openai.com/index/hello-gpt-4o/}.} to generate ten distinct instructions per dataset.

Our instruction-tuning EgoIT dataset can be broadly categorized into three areas: 1) Action Recognition, which focuses on enhancing the identification of hand-object interactions. We utilize the EGTEA~\cite{li2021eye} and Something-Something-V2~\cite{goyal2017something} datasets within the egocentric setting. 2) Question Answering, which aims to improve the ability to answer questions based on video content. We include the EgoTimeQA~\cite{di2024grounded} and OpenEQA~\cite{Majumdar_2024_CVPR} datasets, encompassing both open-ended and close-ended questions. And 3) Captioning, which pays attention to enhancing basic visual description capabilities, utilizing the EgoExoLearn~\cite{huang2024egoexolearn} dataset for this purpose.

%% file: sec/4_method.tex
\section{Method}

Under limited data conditions, bridging the gap between exocentric and egocentric domains is essential for improving the performance of MLLMs on various egocentric tasks. To tackle this challenge, we propose a progressive multimodal training paradigm, as illustrated in Figure~\ref{fig:method}. We describe each step in detail below.

\begin{table*}[t]
    \renewcommand{\arraystretch}{1.0}
    \centering
    \begin{tabular}{c|c|cc}
        \hline
        \multirow{2}{*}{\textbf{Process}} & \multirow{2}{*}{\textbf{Dataset}} & \multicolumn{2}{c}{\textbf{Total}}\\
        & & Exo& Ego \\
        \hline
        \multirow{2}{*}{Initialization}&Panda-70M, WebVid-10M, VIDAL-10M, InternVid-10M&\multirow{2}{*}{103M}&\multirow{2}{*}{3.8M}\\
        &CC-3M, DCI-7.8K, EgoClip-3.8M&\\
        Stage 1& Ego-ExoClip-1.1M&1.1M&-\\
        Stage 2& Ego-ExoClip-1.1M&1.1M&261K\\
        \multirow{2}{*}{Stage 3}& OpenEQA-1.6K, EgoTimeQA-303K, Something-Something-V2-221K&\multirow{2}{*}{-}&\multirow{2}{*}{586K}\\
        &EGTEA-15K, EgoExoLearn-45K&\\
        \hline
    \end{tabular}
    \caption{Overview of the datasets utilized in the multiple training stages, detailing their sources and scales.}

    \label{tab:dataset}
\end{table*}

\begin{table*}[t]
    \renewcommand{\arraystretch}{1.0}
    \centering
    \begin{tabular}{cccccccccccc}
        \hline
        \multirow{2}{*}{\textbf{Method}} & \textbf{ES} & \multicolumn{2}{c}{\textbf{QAEgo4D}} & \multicolumn{2}{c}{\textbf{EgoTaskQA}} &\textbf{CE} &\textbf{EK}&\textbf{EP}&\textbf{VLN-QA}&\multicolumn{2}{c}{\textbf{EgoMCQ}}\\
        & Acc. & Acc.& Acc./Score&  Acc. & Acc.&mAP&mAP& Acc.& Acc.& Acc. & Acc.\\
         \hline
         \multicolumn{12}{c}{\textit{Specific Egocentric Methods}}\\
         \hline
         EgoVLP &- & 22.6& 17.3/1.9& 42.5 & 38.7&25.0&26.0&-&-&-&-\\
         EgoVLPv2&- & 25.8 & 19.7/2.1& 46.3 & 42.3&30.7&29.1&15.3&10.4&72.3&30.6\\
         MFAS&- &37.1 & 23.2/2.2& 48.7 & 45.4&37.4&30.3&17.2&9.5&77.5&34.2\\
         GroundVQA &- & 50.2 & 29.0/2.6& - & -&42.5&-&22.7&17.4&80.3&37.1\\
         LaViLa++&-&-&-&-&-&40.8&-&20.9&11.0&75.4&33.7\\
         \hline
        \multicolumn{12}{c}{\textit{Zero-shot Closed-source MLLMs}}\\
        \hline
        Gemini 1.5-Pro &71.2&-&-&-&-&-&-&-&-&-&-\\
        GPT-4o&72.2&60.3&29.1/3.0&43.2&47.5&59.4&45.3&36.8&34.0&86.6&38.8\\
        \hline
         \multicolumn{12}{c}{\textit{Zero-shot Open-source MLLMs}}\\
         \hline
        LLaMA-VID &38.5&38.6&23.5/2.0&29.0&31.6&31.0&46.1&27.2&32.3&56.5&23.5\\
         Video-LLaVA&38.4&35.2&\textbf{28.7}/2.7&29.8&30.6&43.5&45.2&27.5&34.0&60.9&25.1\\
        VideoLLaMA2&51.7&48.5&25.3/2.6&42.1&42.6&68.1&47.8&29.3&37.5&82.1&32.0\\
        VideoChat2&54.4&45.4&21.9/1.7&42.3&47.4&57.0&42.6&26.1&35.0&85.1&37.6\\
        Qwen2.5-VL&57.6&55.3&26.4/2.8&42.6&45.8&64.2&46.4&32.5&34.3&85.3&38.2\\
         \rowcolor{gray!20} \textbf{Exo2Ego}&\textbf{61.3}&\textbf{62.1}&28.3/2.7&\textbf{44.7}&\textbf{50.3}&\textbf{70.9}&\textbf{49.7}&\textbf{42.7}&\textbf{44.5}&\textbf{88.4}&\textbf{41.2}\\
         \hline
    \end{tabular}
        \caption{Performance comparison with recent state-of-the-art methods methods on eight benchmarks, where ES, CE, EK, and EP refer to EgoSchema, Charades-Ego, EPIC-KITCHENS-100, and EgoPlan. The QAEgo4D dataset is categorized into \textit{close} and \textit{open} question answering, EgoTaskQA includes \textit{direct} and \textit{indirect} subsets, and EgoMCQ contains \textit{inter} and \textit{intra} settings. The best results among open-source MLLMs are highlighted in bold.}

    \label{tab:vqa}
\end{table*}

\subsection{Initialization}
The goal of the initialization is to equip both the exocentric and egocentric visual encoders with foundational capabilities to process their respective data types. We tackle this by training the encoders separately using data from both domains. For the exocentric visual encoder, we utilize a large-scale, web-crawled dataset comprising image-text and video-text pairs from various publicly accessible databases, as summarized in Table~\ref{tab:dataset}. For the egocentric visual encoder, we employ the EgoClip~\cite{lin2022egocentric} dataset, which includes a wide range of first-person daily activities. During this training phase, the parameters of LLM are frozen, and we optimize only the two visual encoders. The trained parameters of visual encoders then serve as initialization weights for the subsequent three-stage process.

\subsection{Stage 1: Demonstrator Self-Preparation}
After initialization, the exocentric visual encoder (the demonstrator) must adapt to the Ego-ExoClip dataset, despite already being proficient at understanding exocentric videos. To achieve this, we implement a self-preparation step to ensure that the exocentric encoder is adequately prepared to transfer its knowledge to the egocentric branch. 
As shown in Figure~\ref{fig:method}, we freeze the LLM parameters and fine-tune the exocentric visual encoder using exocentric clip-text data from our proposed Ego-ExoClip dataset. For supervision, we employ Vision-grounded Text Generation (VTG) loss to ensure data accuracy. 

\subsection{Stage 2: Demonstrator-Learner Guidance}
In the second stage, we aim to establish a robust correlation between the exocentric and egocentric data, effectively mapping the learner to the demonstrator. This mapping is essential, as exocentric knowledge can significantly enhance the comprehension of egocentric videos, given that human behaviors are largely perspective-invariant and independent of camera viewpoint. 
For this stage, we continue to adopt Ego-ExoClip as input data. With the exocentric visual encoder sufficiently trained, we freeze it to maintain its integrity. We also freeze the LLM while fine-tuning the egocentric visual encoder and the mapping functions $F$ and $G$. The mapping $F: X \rightarrow Y$ is inherently under-constrained, prompting us to couple it with an inverse mapping $G: Y \rightarrow X$, where $X$ and $Y$ denote the egocentric and exocentric domains, respectively. 
Both mappings are treated as bijections, and we enforce this structural assumption by training them simultaneously while
applying a Cycle Consistency Loss (CCL), defined as: 

\begin{equation}
\begin{aligned}
   \mathcal{L}_{\text{CCL}}(F, G) = & \ \mathbb{E}_x \left[ \|G(F(x)) - x\|_1 \right] \\
+ & \ \mathbb{E}_y \left[ \|F(G(y)) - y\|_1 \right],
\end{aligned}
\end{equation}
where $x \in X$, $y \in Y$, and $\|\|_1$ denotes the L1 norm. The CCL encourages both forward cycle consistency, i.e., $x \rightarrow F(x) \rightarrow  G(F(x)) \approx x$, and backward cycle consistency, i.e., $y \rightarrow G(y) \rightarrow  F(G(y)) \approx y$.
Additionally, we introduce the Kullback–Leibler (KL) divergence between the exocentric sample $y$ and the estimated exocentric sample $\hat{y}=F(x)$, aligning real and estimated exocentric feature distributions. The concatenated guidance from the demonstrator and the learner's own information is then fed into the LLM.

\subsection{Stage 3: Learner Self-Practice}
In the final stage, we focus on enhancing the instruction-following capabilities of MLLMs to address diverse egocentric downstream tasks. To achieve this, we employ the instruction-tuning data EgoIT. By concatenating the representations of the egocentric sample $x$ and its corresponding mapped exocentric sample $F(x)$, we create a comprehensive semantic input for the LLM. To ensure the LLM aligns its responses more effectively with task instructions, we apply Low-Rank Adaptation (LoRA) to the frozen LLM, tuning it alongside the visual encoder and the mapping function $F$ using VTG loss.

%% file: sec/5_experiment.tex
\section{Experiment}

\subsection{Implementation Details}

\textbf{Datasets.} The datasets employed in each stage of training are summarized in Table~\ref{tab:dataset}, showcasing a substantial amount of vision-text data across various types. After training, we evaluated our model on multiple benchmarks, which includes eight tasks with multiple configurations:  EgoSchema~\cite{mangalam2023egoschema} for reasoning, QAEgo4D~\cite{barmann2022did} for episodic memory, Charades-Ego~\cite{sigurdsson2018charades} for action recognition, EPIC-KITCHENS-100~\cite{damen2022rescaling} for multi-instance retrieval, VLN-QA~\cite{krantz2020beyond} for egocentric navigation, EgoPlan~\cite{chen2023egoplan} for planning, EgoMCQ~\cite{lin2022egocentric} for alignment, and EgoTaskQA~\cite{jia2022egotaskqa} for hybrid tasks. 

\noindent\textbf{Model.} Our model is built upon the VideoLLaMA2~\cite{cheng2024videollama} baseline. We consistently utilized the CLIP-Large-336 as the visual encoders and Mistral-7B-Instruct as the LLM. The mapping functions $F$ and $G$ consist of nine ResNet blocks between the down-sampling and up-sampling operations. In Stage 3, we incorporated LoRA into the LLM, with a rank set to 128, an alpha value of 256, and a dropout rate of 0.1. All experiments are conducted using the PyTorch framework on 16 A800 GPUs. 

\noindent\textbf{Baselines.} We selected three categories of methods as comparative baselines, including specific approaches tailored for the egocentric domain (e.g., GroundVQA~\cite{di2024grounded} and LaViLa++~\cite{xu2025egocentric}), closed-source MLLMs (e.g., GPT-4o), and open-source MLLMs (e.g., VideoLLaMA2~\cite{cheng2024videollama} and Qwen2.5-VL~\cite{bai2025qwen2}). Except for LLaMA-VID~\cite{li2023llama} and Video-LLaVA~\cite{lin2023video}, which use fixed input settings of 1 fps and 8 frames respectively, all other models adopt a 16-frame configuration to ensure a fair comparison.

\subsection{Quantitative Evaluation}
Tables~\ref{tab:vqa} presents the experimental results with different superior methods across various tasks and settings, which can be analyzed as follows: 
\begin{itemize}
     \item The experimental results on the reasoning and episodic memory tasks (i.e., the first three benchmarks) indicate that, in a zero-shot setting, MLLMs achieve performance comparable to specialized egocentric methods, highlighting their strong generalization capability across data domains. Notably, our Exo2Ego method significantly surpasses all open-source MLLMs and egocentric-specific methods under nearly all settings across the three egocentric evaluation benchmarks. 
    \item The results on the fourth and fifth benchmarks present zero-shot performace on action recognition (Charades-Ego) and multi-instance retrieval (EPIC-KITCHENS-100). Our Exo2Ego method demonstrates consistent improvements over all baselines across all metrics for both tasks, validating its effectiveness in understanding the details of hand-object interactions.
    \item Planning and navigation are critical tasks in the domain of egocentric understanding. The sixth and seventh columns summarize our method's performance against existing baselines. The results reveal that open-source MLLMs exhibit accuracy rates concentrated between 26.0\% and 32.5\% on EgoPlan, highlighting their limitations for planning.
    In contrast, our method significantly outperforms all competitors, including GPT-4o (5.9\% and 10.5\% absolute gains), demonstrating its strong capability to effectively tackle diverse egocentric tasks. 
    \item Video-text matching, a fundamental alignment task, is also crucial for comprehensive evaluation. As shown in the final column, our Exo2Ego approach consistently achieves the best performance across both settings, further demonstrating its superiority in generating accurate visual descriptions.
\end{itemize}

\subsection{Ablation Study}
In this section, we conducted a comprehensive analysis of how our model architecture, parameter update, training data, and training strategy impact overall performance. 

\noindent\textbf{Model Architecture.} As summarized in Table~\ref{tab:ab_ma}, we investigated the impact of the mapping functions, LoRA, and loss functions on model performance. Comparing results 1 and 2, we observed that the introduction of LoRA significantly improves instruction-following capabilities, leading to enhanced performance across various tasks. This highlights LoRA's effectiveness in refining the model's ability to follow complex instructions. 
The third experiment utilizes only forward cycle consistency, resulting in a performance decline and highlighting the importance of the full bidirectional loss design.
In the comparison between results 1 and 4, we found that the mapping $G$ and CCL loss play a critical role in establishing an effective mapping between the egocentric and exocentric domains, 
thereby improving performance on downstream egocentric tasks. 
In the fifth experiment, we removed the KL loss, which incorporates guidance from exocentric data. The performance gap between this result and result 1 reinforces the importance of exocentric knowledge in enhancing egocentric video understanding.
In the sixth experiment, we replaced the existing mapping functions with fully connected networks, which led to a slight performance decline. This suggests that more complex network architectures are capable of learning improved mapping transformations.

\begin{table}[t]
    \renewcommand{\arraystretch}{1.0}
    \centering
    \begin{tabular}{cccccc|c}
        \hline
        \multirow{2}{*}{\textbf{Num}}&\multicolumn{2}{c}{\textbf{Mapping}}& \multirow{2}{*}{\textbf{LoRA}}&\multicolumn{2}{c|}{\textbf{Loss}}&\multirow{2}{*}{\textbf{Avg}} \\
        &$F$ & $G$ & & KL& CCL &\\
         \hline
         \rowcolor{gray!20} 1&\ding{51}&\ding{51}&\ding{51}&\ding{51}&\ding{51}&\textbf{55.6}\\     2&\ding{51}&\ding{51}&\ding{55}&\ding{51}&\ding{51}&53.2~\textcolor{red}{$\downarrow$}\\
         3&\ding{51}&\ding{55}&\ding{51}&\ding{51}&\ding{233}&54.9~\textcolor{red}{$\downarrow$}\\
         4&\ding{51}&\ding{55}&\ding{51}&\ding{51}&\ding{55}&54.4~\textcolor{red}{$\downarrow$}\\
         5&\ding{51}&\ding{51}&\ding{51}&\ding{55}&\ding{51}&51.4~\textcolor{red}{$\downarrow$}\\
         6&\ding{80} &\ding{80}&\ding{51}&\ding{51}&\ding{51}&54.7~\textcolor{red}{$\downarrow$}\\
         \hline
    \end{tabular}
        \caption{Performance comparison of different model architecture. The LLM is fixed at Mistral-7B-Instruct-v0.2, ``\ding{80}" denotes replacing the existing modules with fully connected networks, and ``\ding{233}" represents removing backward cycle consistency, using only the forward direction.
        ``Avg'' is the average performance across all metrics in Figure~\ref{fig:intro}.}

    \label{tab:ab_ma}
\end{table}

\begin{table}[t]
    \renewcommand{\arraystretch}{1.0}
    \centering
    \begin{tabular}{cccc|c}
        \hline
        \multirow{2}{*}{\textbf{Num}}&\multicolumn{2}{c}{\textbf{Stage 2}}& \textbf{Stage 3}&\multirow{2}{*}{\textbf{Avg}} \\
         &Exo ViEnc & Ego ViEnc & Ego ViEnc&  \\
         \hline
         \rowcolor{gray!20} 1&\scalebox{0.8}{\faSnowflake} &
         \scalebox{0.8}{\faFire}&\scalebox{0.8}{\faFire}&\textbf{55.6}\\
         2&\scalebox{0.8}{\faFire} &
         \scalebox{0.8}{\faFire}&\scalebox{0.8}{\faFire}&54.9~\textcolor{red}{$\downarrow$}\\
         3&\scalebox{0.8}{\faSnowflake} &
         \scalebox{0.8}{\faFire}&\scalebox{0.8}{\faSnowflake}&52.5~\textcolor{red}{$\downarrow$}\\

         \hline
    \end{tabular}
        \caption{Effects of different parameter update settings.}

    \label{tab:ab_tm}
\end{table}

\begin{table}[t]
    \renewcommand{\arraystretch}{1.0}
    \centering
    \begin{tabular}{ccc|c}
        \hline
         \textbf{EgoClip}& \textbf{Ego-ExoClip}&\textbf{EgoIT}&{\textbf{Avg}}\\
         \hline
        \ding{55}& \ding{55} &\ding{55}&38.9\\
        \ding{51}& \ding{55} &\ding{55}&45.2~\textcolor{OliveGreen}{$\uparrow$}\\
         \ding{51}& \ding{51} &\ding{55}&47.8~\textcolor{OliveGreen}{$\uparrow$}\\
         \ding{51}& \ding{51} &\ding{51}&49.7~\textcolor{OliveGreen}{$\uparrow$}\\
         \rowcolor{gray!20}\ding{51}& \ding{51} &\ding{51}&\textbf{55.6 (Ours)}\\
         \hline
    \end{tabular}
        \caption{Effect comparison across different datasets. The results in the first four rows are from VideoLLaMA2, using the same backbone as our method.}

    \label{tab:ab_id}
\end{table}

\noindent\textbf{Parameter Update.} The choice of visual encoders plays a crucial role in determining overall model performance, as detailed in Table~\ref{tab:ab_tm}. In the second experiment, we simultaneously fine-tuned both the exocentric and egocentric visual encoders during Stage 2, allowing the teacher and student to learn from each other. 
Comparing results 1 and 2, we observed that fine-tuning both encoders together may lead to the establishment of a neutral mapping between the domains, which ultimately results in poorer downstream performance. The gap between results 1 and 3 emphasizes the importance of maintaining more learnable parameters for visual adaptation during instruction fine-tuning.

\noindent\textbf{Training Data.} To assess the influence of training data on the model performance, we conducted experiments using VideoLLaMA2~\cite{cheng2024videollama} as a baseline, as summarized in Table~\ref{tab:ab_id}. Notably, VideoLLaMA2 adopts the same backbone architecture as our framework, including visual encoder and LLM.
In the second row of results, we added the EgoClip dataset, which is aligned with the egocentric domain. The improved performance suggests that training on data closely related to the downstream tasks leads to better domain knowledge. 
Compared to result 2, result 3 incorporates the Ego-ExoClip data and achieves further performance improvements. This validates the effectiveness of our constructed dataset.
The fourth result indicates a significant improvement over the third result, demonstrating that our egocentric instruction-tuning data EgoIT enhances adaptation to downstream egocentric tasks. 
The performance gap between the last two rows indicates that, even when using the same datasets, our framework still exhibits a significant advantage. This demonstrates the superiority of our dual-encoder architecture and transfer-based training strategy.

\begin{figure}[t]
  \centering
   \includegraphics[width=0.95\linewidth]{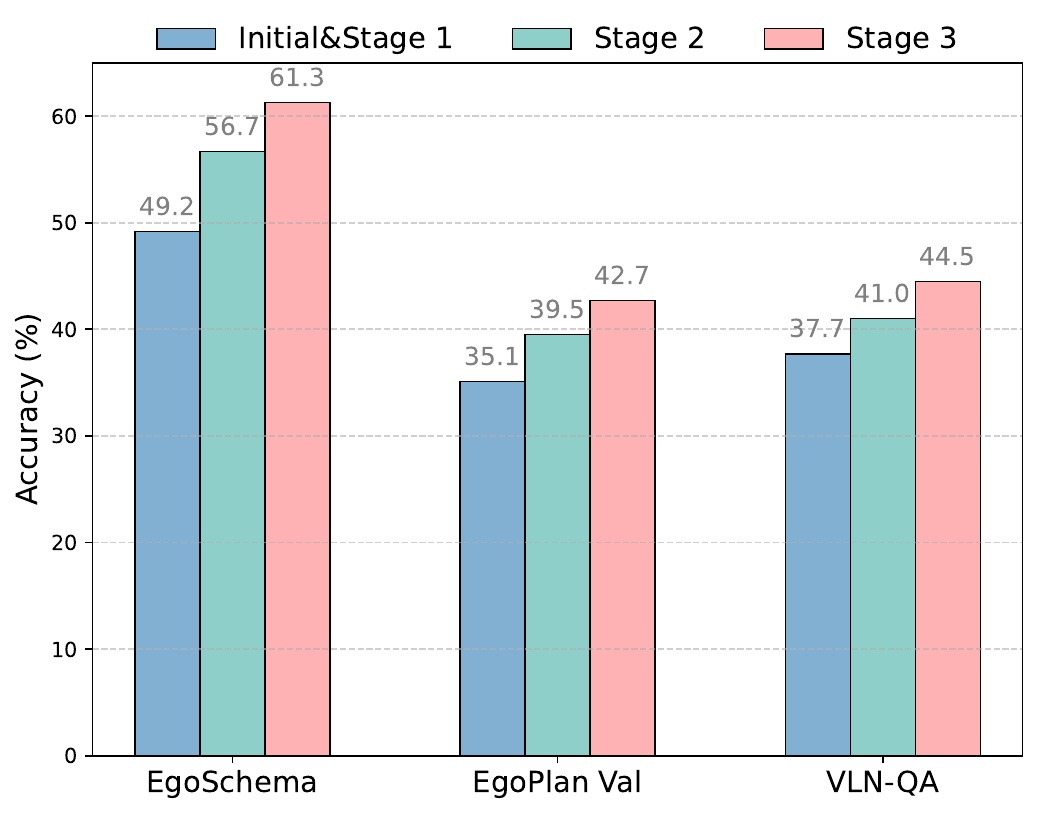}
   \caption{Accuracy comparison across different stages.}
   \label{fig:bar}
\end{figure}

\noindent\textbf{Training Strategy.} To quantify the contribution of each training stage, we evaluated each stage separately and the results are shown in Figure~\ref{fig:bar}. The results show that pre-training establishes a foundation for basic egocentric video comprehension, while mapping learning in Stage 2 effectively aligns cross-domain data, yielding competitive performance. Moreover, the enhanced instruction-following capability achieved in Stage 3 further improves performance on downstream tasks. These results substantiate the efficacy of our three-stage training pipeline.

\begin{figure}[t]
  \centering
   \includegraphics[width=0.97\linewidth]{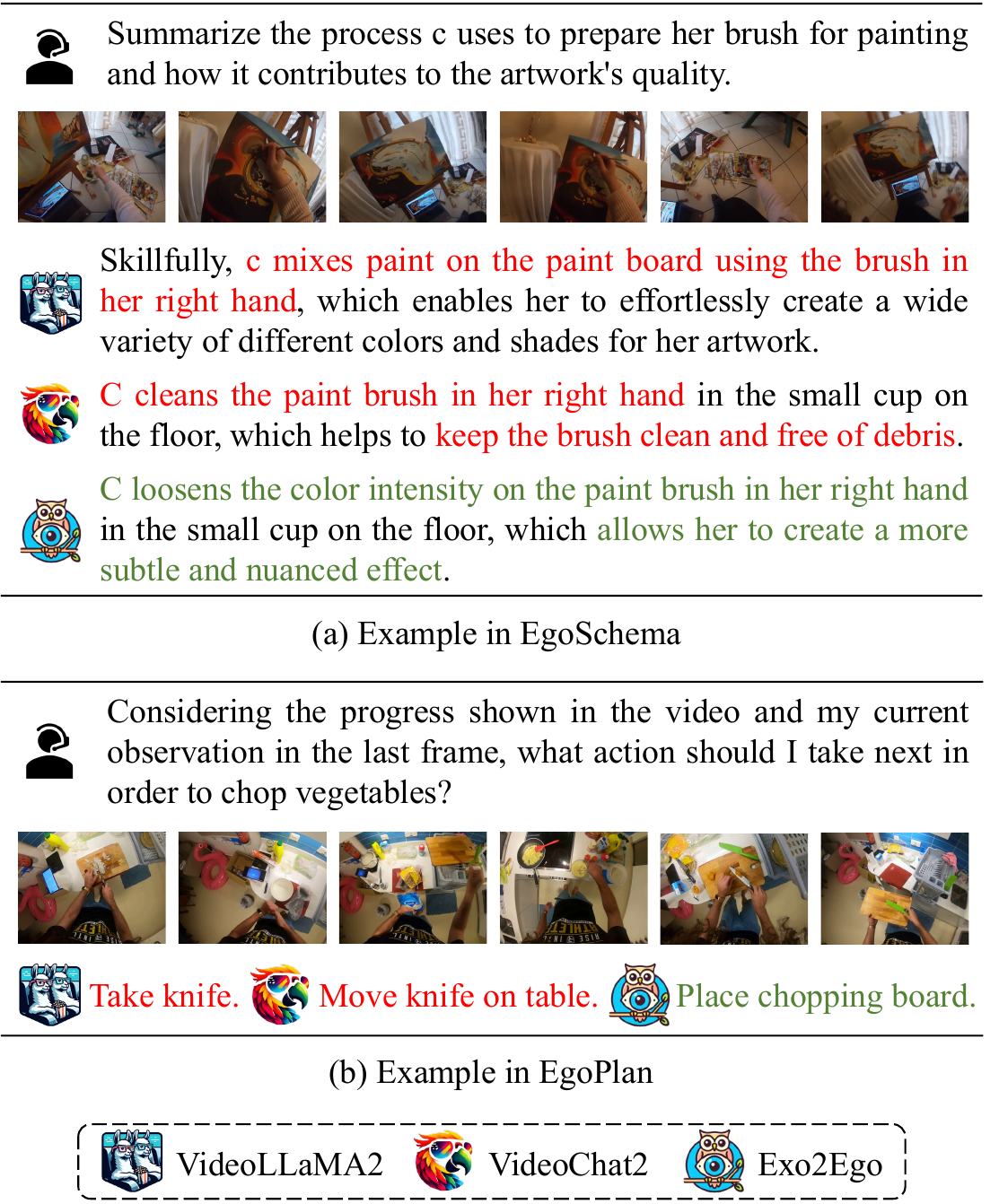}
  \caption{Two examples from different benchmarks.}
   \label{fig:vis}
\end{figure}

\subsection{Qualitative Comparison}
To visually compare the performance of different MLLMs, we presented several qualitative examples in Figure~\ref{fig:vis}. In Figure~\ref{fig:vis}(a), both baselines struggle with accurately interpreting the video content. VideoLLaMA2 erroneously assumes the camera-wearer is mixing paint with a brush, while VideoChat2 mistakenly believes the brush is being cleaned. In contrast, our Exo2Ego method correctly identifies that the camera-wearer is diluting the color on the brush, showcasing its superior understanding and inference capabilities for visual details.
In the second example from EgoPlan dataset, which evaluates planning abilities, our model demonstrates a clear understanding of the temporal relationships between actions. Unlike the two baselines, which provide incorrect predictions, our model accurately identifies the correct sequence of actions required to achieve the goal. 

%% file: sec/6_conclusion.tex
\section{Conclusion}
In this paper, we propose Exo2Ego, a unified framework that significantly advances egocentric video understanding by cost-effectively transferring rich exocentric knowledge from MLLMs. To enable robust cross-view representation learning, we introduce Ego-ExoClip, a large-scale dataset comprising 1.1 million synchronized egocentric-exocentric clip-text pairs. Additionally, we curate EgoIT, an egocentric instruction-tuning dataset designed to further enhance instruction-following capabilities.
By leveraging a novel exocentric-to-egocentric migration strategy and a progressive three-stage training pipeline, Exo2Ego effectively bridges the domain gap and promotes egocentric self-consistency. Extensive experiments on multiple benchmarks demonstrate that Exo2Ego consistently achieves substantial performance improvements.



%% file: sec/X_suppl.tex

\twocolumn[
  \begin{center}
    \LARGE \textbf{— Supplementary Material —}
  \end{center}
  \vspace{2em}
]

\setcounter{page}{1}
\setcounter{section}{0}
\renewcommand{\thesection}{\Alph{section}}


\section{Ego-ExoClip Details}
\label{eecd}
\subsection{Dataset Comparison}\label{adc}
In Table~\ref{tab:sm_rd}, we compare our proposed Ego-ExoClip dataset with existing egocentric-exocentric datasets, including EgoExoLearn~\cite{huang2024egoexolearn}, Ego-Exo4D~\cite{grauman2024ego}, and EgoExo-Fitness~\cite{li2024egoexo}, all published in 2024. 
Ego-ExoClip preserves the strengths of Ego-exo4D, such as diverse activity scenarios, 
richly synchronized exocentric perspectives, and a large-scale dataset.
Moreover, compared to Ego-Exo4D, Ego-ExoClip extends timestamp-level narrations to the clip level, enabling a broader range of pre-training tasks. Therefore, 
Ego-ExoClip is the largest and most diverse clip-text dataset currently available, offering synchronized egocentric-exocentric perspectives.

\begin{table*}[t]

    \renewcommand{\arraystretch}{1.1}
    \centering
    \begin{tabular}{c|ccccccc}
        \hline
        \textbf{Dataset} &\textbf{Year}&\textbf{Scenario}&\textbf{Ego\&Exo?}&\textbf{Exo Num}& \textbf{Narration}&\textbf{Clip-Text}&\textbf{Unique Hours}\\
         \hline
         CharadesEgo&2018&Indoor Daily&Paired&1&\ding{51}&\ding{51}&34\\
         LEMMA&2020&Kitchen, Living &Paired&1&\ding{55}&\ding{55}&10\\
         HOMAGE&2021&Indoor Daily&Paired&2-5&\ding{55}&\ding{55}&25\\
         Assembly101&2022&Desk&Paired&8&\ding{55}&\ding{55}&42\\
         EgoExoLearn&2024&Kitchen, Lab&Unpaired&-&\ding{51}&\ding{51}&120\\
         Ego-Exo4D&2024&Diverse&Paired&4-5&\ding{51}&\ding{55}&221\\
         EgoExo-Fitness&2024&Fitness &Paired&3 &\ding{51} &\ding{51} &32\\
         \hline
         Ego-ExoClip&-&Diverse&Paired&4-5&\ding{51}&\ding{51}&116\\
         \hline
    \end{tabular}
        \caption{Comparison of related egocentric-exocentric datasets. ``Ego\&Exo?" refers to whether egocentric and exocentric perspectives are synchronized. ``Exo Num" represents the number of fixed cameras capturing exocentric perspectives simultaneously. 
    ``Narration" refers to the availability of detailed behavioral descriptions, beyond simple action labels. 
    ``Clip-Text" indicates whether the dataset provides clip-text pairs 
    for pre-training.}
    \label{tab:sm_rd}
\end{table*}

\subsection{Clip Analysis}\label{aca}
\begin{figure}[t]
  \centering
   \includegraphics[width=\linewidth]{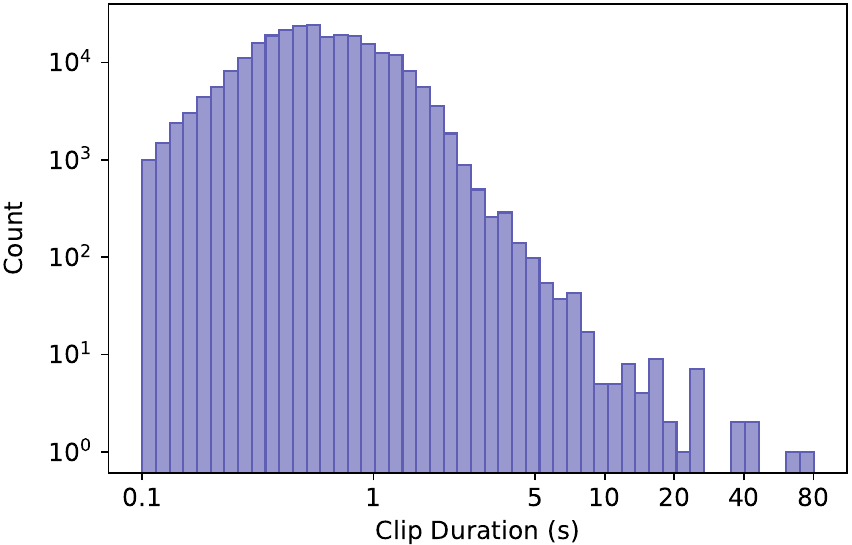}
 \caption{Distribution of clip durations in Ego-ExoClip.}
   \label{fig:sm_do}
\end{figure}

Figure~\ref{fig:sm_do}
shows the distribution of clip durations in Ego-ExoClip, which includes a total of 261.3K clips. The average clip duration is 0.68 seconds, with a standard deviation of 0.57 seconds. Notably, 82.0\% of the clips are shorter than 1.0 seconds, reflecting the abundance of atomic, instantaneous actions densely labeled by Ego-Exo4D. The longest clip lasts 70.43 seconds, corresponding to the narration ``C starts playing the violin'' in the \textbf{Music} scenario, while the shortest clip captures ``C steps her left foot forward'' in the \textbf{Dance} scenario.
\subsection{Narration Analysis}\label{ana}
\begin{figure}[t]
  \centering
   \includegraphics[width=\linewidth]{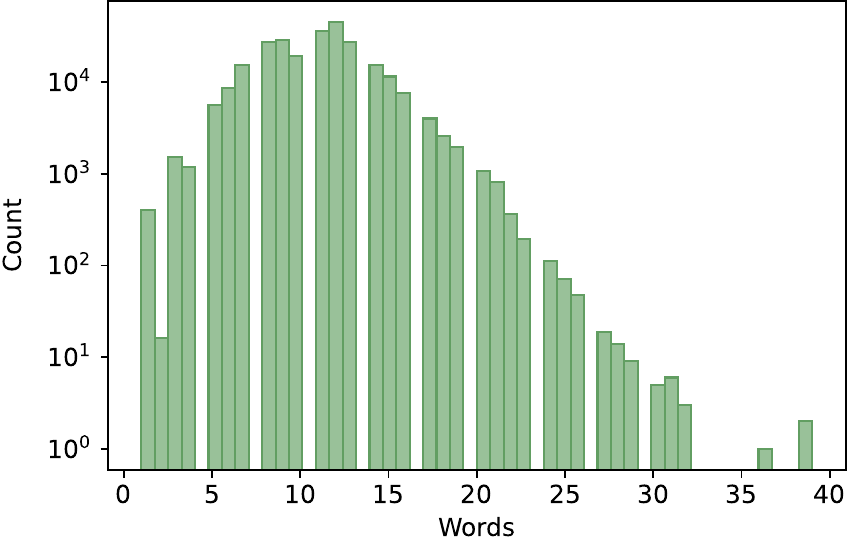}
  \caption{Distribution of narration word lengths in Ego-ExoClip.}
   \label{fig:sm_wl}
\end{figure}

Figure~\ref{fig:sm_wl} displays the word length distribution for the 261.3K narrations in Ego-ExoClip. The average narration length is 10.98 words, with a standard deviation of 3.16 words. Notably, 65.5\% of the narrations contain between 10 and 20 words, indicating that the narrations in Ego-EcoClip are generally detailed and descriptive. The longest narration, found in the \textbf{Health} scenario, is ``C throws out the covid 19 test tube seal, the covid 19 test box, the torn part of the covid 19 test device pack, and the covid 19 test instruction sheet into the waste bag with his right hand".

\subsection{Others}\label{ao}
\noindent \textbf{Scenario Diversity.} Figure~\ref{fig:sm_sd} presents the scenario distribution of the videos in Ego-ExoClip, which includes a total of 2,925 groups across eight specific human daily scenarios, such as Cooking, Health, and Bike Repair.
Among all scenarios, \textbf{Rock Climbing} represents the highest proportion at 27.8\%, while \textbf{Music} accounts for the lowest at 5.5\%. 
Each scenario also has a wide range of diversity, with participants from multiple cities featured in Ego-Exo4D. For example, the \textbf{Health} scenario includes videos from Singapore and Tokyo, while \textbf{Cooking} is captured in Vancouver and Minneapolis. 

Each scenario covers a variety of specific tasks, such as Cooking (``cooking an omelet", ``cooking tomato\&eggs", and ``cooking noodles"), Bike Repair (``remove a wheel", ``adjust a rear derailueur", and ``clean and lubricate the chain"), Music (``playing violin", ``playing guitar", and ``playing piano"). Due to the large number of tasks, we only display the top-50 in Figure~\ref{fig:sm_td}. 
Notably, this distribution is long-tailed, where the most common 
task, ``basketball drills-reverse layup" includes 312 sets of videos (10.7\%), whereas the 
least common 
task, ``playing guitar-scales and arpeggios" contains just 21 sets.

\begin{figure}[t]
  \centering
   \includegraphics[width=\linewidth]{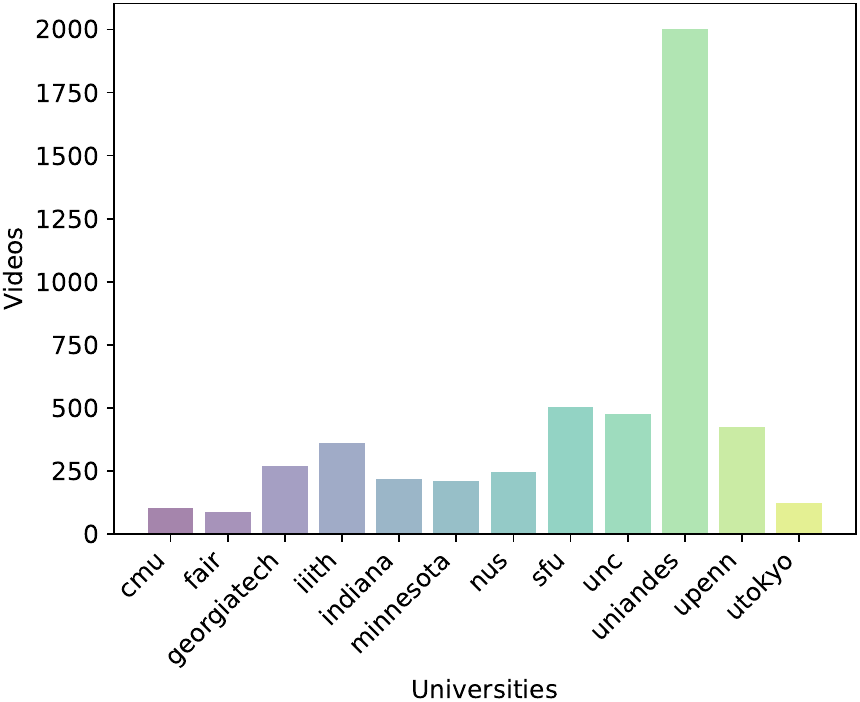}
   \caption{Institution distribution of Ego-ExoClip.}
   \label{fig:sm_gd}
\end{figure}

\begin{figure*}[t]
  \centering
   \includegraphics[width=\linewidth]{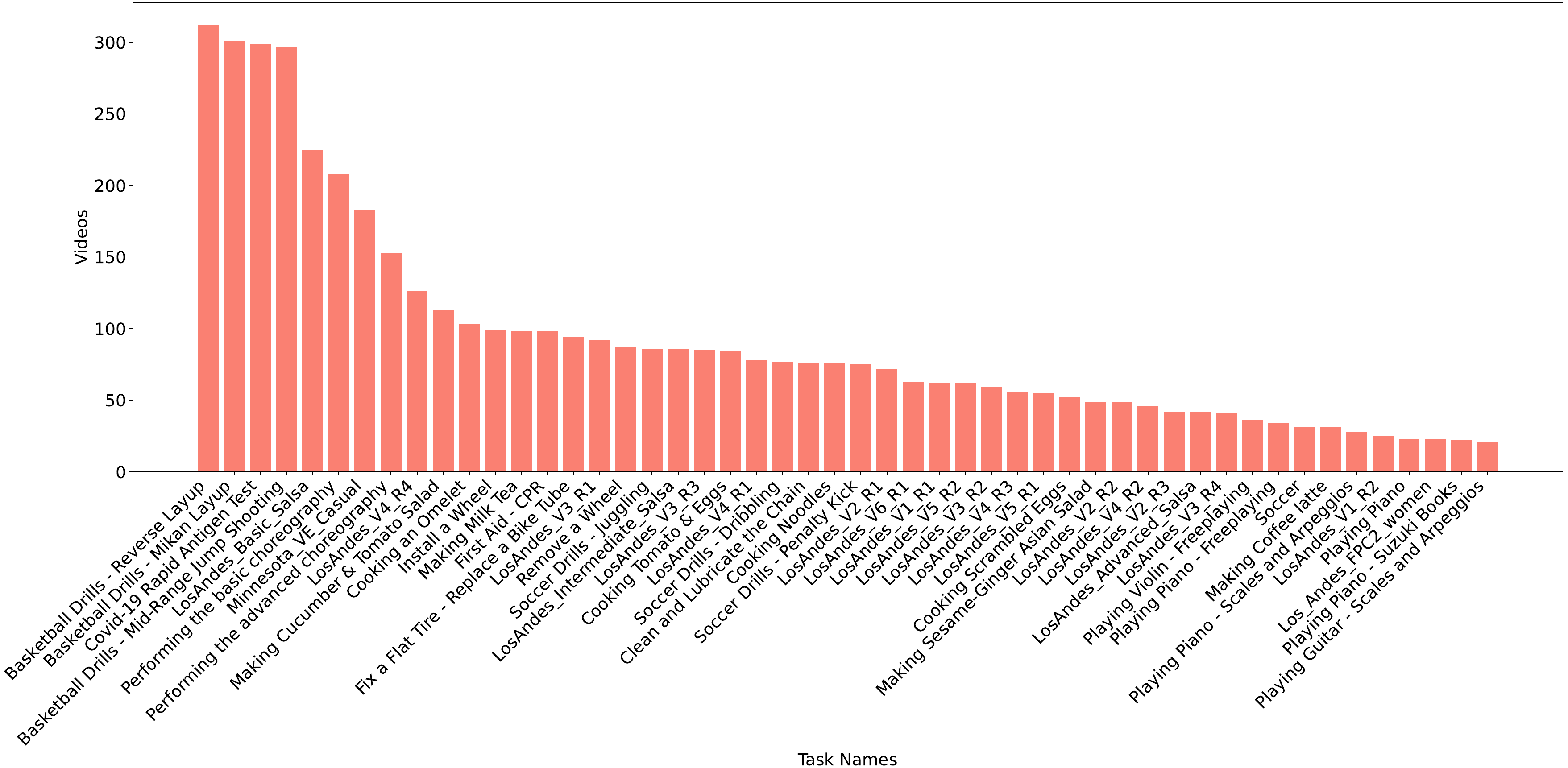}
   \caption{Task distribution of Ego-ExoClip.}
   \label{fig:sm_td}
\end{figure*}

\begin{figure}[t]
  \centering
   \includegraphics[width=\linewidth]{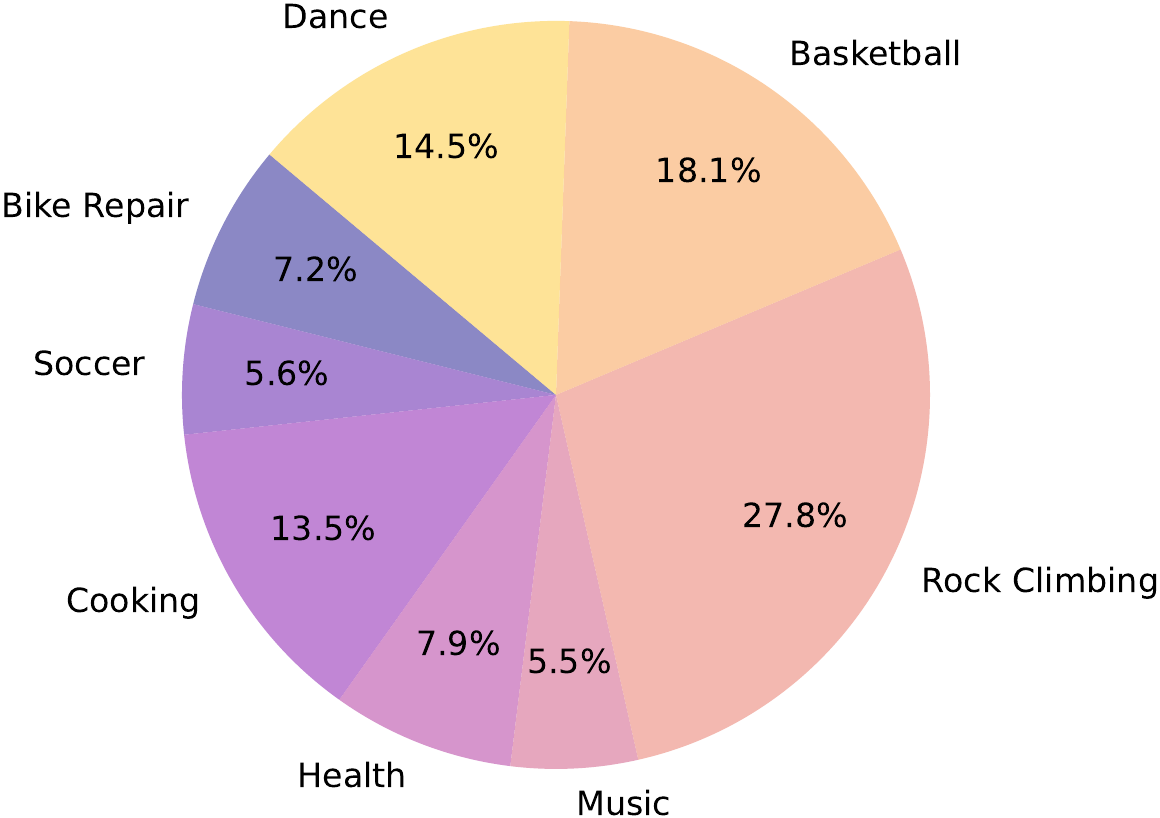}
   \caption{Scenario distribution of Ego-ExoClip.}
   \label{fig:sm_sd}
\end{figure}

\noindent \textbf{Geographic Diversity.} Figure~\ref{fig:sm_gd} shows the regional distribution of the videos in Ego-ExoClip, which includes a total of 2,925 groups. The dataset covers contributions from 
12 institutions across 6 different countries: India, Singapore, Japan, USA, Canada, and Colombia. 
This broad coverage allows our pre-training dataset to effectively inherit the rich geographic and participant diversity of Ego-Exo4D. For more detailed information, please refer to the original Ego-Exo4D paper~\cite{grauman2024ego}.

\section{EgoIT Details}
\label{egoitd}
\begin{figure}[t]
  \centering
   \includegraphics[width=\linewidth]{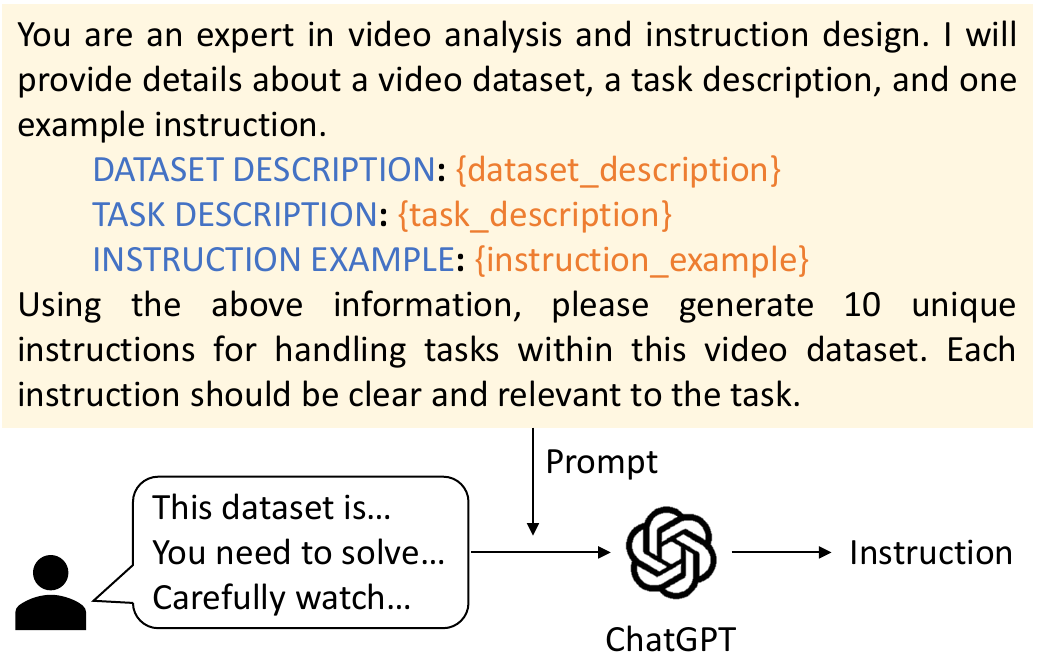}
   \caption{Prompt used for the instruction generation process in the EgoIT dataset.}
   \label{fig:sm_ig}
\end{figure}

Figure~\ref{fig:sm_ig} shows the prompt used to generate instructions for different types of datasets in EgoIT. The descriptions differ across datasets and tasks, leading to the generation of diverse instructions, as detailed in Table~\ref{tab:sm_dd}. The descriptions for different tasks are as follows:  \textbf{Action Recognition} is the task of identifying and classifying human actions within video sequences; 
\textbf{Question Answering} is the task of analyzing video content to generate accurate answers to questions based on the visual information within the video;
\textbf{Captioning} is the task of generating descriptive textual summaries that capture the key events and actions occurring in the video.

\begin{table*}[!t]
    \renewcommand{\arraystretch}{1.1}
    \centering
    \begin{tabular}{p{6cm}p{10cm}}
        \hline
         \textbf{Dataset Description}& \textbf{Generated Instruction}\\
         \hline
        \multirow{9}{*}{\parbox{6cm}{\textbf{EGTEA} (EGTEA Gaze+) is a large-scale FPV dataset with 28 hours of cooking activities from 32 subjects, annotated with 10,325 instances of fine-grained actions and hand masks. It includes human-object interactions like ``Cut bell pepper" and ``Pour condiment into salad".}} & \textbf{1.}Observe the video closely, noting the interactions between the person and the objects around them. Identify the action being performed and categorize it accordingly.\\
        &\textbf{2.}Watch the video and take note of the person's body posture and movement. Look for specific hand gestures or object interactions that help define the action, then classify it.\\
        &\textbf{3.}Carefully observe the sequence of events in the video, noting how the person moves between actions and objects. Determine the most likely action based on your observations.\\
        \hline
        \multirow{9}{*}{\parbox{6cm}{\textbf{Something-Something-V2} dataset consists of labeled video clips of humans performing basic actions with everyday objects, created by crowd workers. It enables machine learning models to learn fine-grained understanding of physical world actions.}}& \textbf{1.}Carefully observe the video and identify key transitions in actions. Pay attention to how the objects are handled and what role they play in the sequence of movements, ensuring the correct action is selected.\\
        &\textbf{2.}Focus on the timing and coordination of movements in the video. Pay special attention to subtle differences in object manipulation and action flow, then select the most fitting classification.\\
        &\textbf{3.}Concentrate on the start, middle, and end of the action sequence, analyzing how the person's interaction with objects evolves. Generate the answer that accurately summarizes the observed action.\\
        \hline
        \multirow{9}{*}{\parbox{6cm}{\textbf{EgoTimeQA} is a dataset containing 303K question-answer pairs sourced from 5,389 video clips. It offers two settings: open-ended QA for free-form responses and close-ended QA for specific, constrained answers.}}&\textbf{1.}Pay attention to the starting and ending points of actions in the video, as well as any transitions or object states. Choose the answer that aligns with the observed progression of events.\\
        &\textbf{2.}Observe the timing and coordination between actions or events in the video, as well as the role each character plays. Generate the answer that accurately reflects these relationships.\\
        &\textbf{3.}Focus on key visual cues, such as hand movements, facial expressions, and object interactions, to understand the context of the action. Select the answer that best matches these details.\\
        \hline
        \multirow{9}{*}{\parbox{6cm}{\textbf{OpenEQA} is the first open-vocabulary benchmark dataset for Embodied Question Answering, supporting episodic memory and active exploration tasks. It includes over 1,600 human-generated questions from more than 180 real-world environments.}}&\textbf{1.}Pay attention to shifts in camera angles or perspectives, as these may draw attention to essential elements in the video that are important for answering the question.\\
        &\textbf{2.}Watch the entire video closely, focusing on any interactions between objects and people. Identify details that could help in forming a correct answer to the question.\\
        &\textbf{3.}Pay close attention to recurring objects and characters throughout the video. Track their involvement and interactions to accurately address the question asked.\\
        \hline
        \multirow{9}{*}{\parbox{6cm}{\textbf{EgoExoLearn}  dataset is designed to simulate the process by which humans learn to follow demonstrations, providing natural annotations for each video clip. It includes footage captured in both everyday life settings and controlled laboratory environments.}}&\textbf{1.}Watch the video attentively, focusing on the main actions and interactions between individuals and objects. Generate a caption that best summarizes these core elements.\\
        &\textbf{2.}Focus on the sequence and timing of events, noting any changes in action or position. Generate a description that captures the progression of events in the most accurate way.\\
        &\textbf{3.}Pay attention to recurring patterns in the video, such as repeated gestures or movements. Generate a summary that accurately reflects these patterns and their impact on the scene.\\
         \hline
    \end{tabular}
       \caption{Descriptions of different datasets in EgoIT and examples of the corresponding generated instructions (showing 3 examples).}

    \label{tab:sm_dd}
\end{table*}

\section{Benchmark Details}
\label{egobenchd}
\subsection{Evaluation Dataset}\label{cdc}
\subsubsection{EgoSchema}
EgoSchema~\cite{mangalam2023egoschema} is a comprehensive dataset and benchmark designed to evaluate the ability of modern vision and language models to understand extended video content. Based on the Ego4D~\cite{grauman2022ego4d} dataset, EgoSchema contains over 5,000 carefully curated multiple-choice questions and answers derived from more than 250 hours of real-life video footage capturing a wide range of natural human activities and behaviors. Each question requires selecting the correct answer from five options after viewing a three-minute video segment, thus challenging models to perform detailed video-based reasoning. 

Since the format of EgoSchema aligns with our requirements, we use it directly as a reasoning evaluation task.

\subsubsection{QAEgo4D}
QAEgo4D~\cite{barmann2022did} is a video question-answering dataset for episodic memory tasks, 
designed to evaluate video comprehension and reasoning capabilities in modern vision-language systems. Sourced from Ego4D~\cite{grauman2022ego4d}, it features human-annotated question-answer pairs, where each sample contains a video, a question, an answer, and the specific temporal window where the answer can be found. The dataset focuses on egocentric videos depicting natural human activities, with an average video length of 310.0 seconds in testing data. By using overlapping text descriptions for consistency, QAEgo4D serves as a valuable resource for advancing research in episodic memory and video question answering.

We use this dataset to support the episodic memory task. For the closed-set QA, we use the testing data processed by Llama2-13B\footnote{\url{https://huggingface.co/meta-llama/Llama-2-13b}.} from~\cite{di2024grounded}, while for the open-set QA, we use the original testing data from~\cite{barmann2022did}.

\begin{table*}[!t]
    \renewcommand{\arraystretch}{1.0}
    \centering
    \begin{tabular}{ccp{10cm}}
        \hline
        \textbf{Task} &\textbf{Benchmark}& \textbf{Example}\\
         \hline
         \multirow{11}{*}{Reasoning}& \multirow{11}{*}{EgoSchema} & \textit{Based on the video, what can you infer about the purpose or function of the sponges and sisal ropes within the context of the construction site work?}\\
         & & (A) sponges and sisal ropes both functioned as alternative communication tools for construction workers on site. (B) sponges and sisal ropes aided in maintaining tools and equipment at construction sites. (C) sponges were used for cleaning purposes, and sisal ropes served to wipe the walls. (D) both sponges and sisal ropes were used to create distinctive markings on surfaces, delineating work areas at the construction site. (E) sponges were used to soak excess water, while sisal ropes functioned as makeshift barriers for specific areas at the site.\\ 
         \hline
         \multirow{3}{*}{Episodic Memory}& \multirow{3}{*}{QAEgo4D} & \textit{Where did I put the chopsticks?}\\
         & & (A) right of the stove. (B) left of the sink. (C) on the table.\\
         & & (D) in the drawer.\\

         \hline
         \multirow{4}{*}{Mixture}& \multirow{4}{*}{EgoTaskQA} & \textit{If I am not pour from something into something, what remaining actions is executable?}\\
         & & (A) open microwave. (B) close microwave. (C) point to microwave.\\
         & & (D) turn off microwave. (E) turn on microwave.\\ 
         \hline
         \multirow{3}{*}{Action Recognition}& \multirow{3}{*}{Charades-Ego} & \textit{What action am I currently performing?}\\
         & & (A) lying on the floor. (B) sitting on a table. (C) lying on a sofa/couch.\\
         & & (D) lying on a bed. (E) sitting in a bed.\\ 
         \hline
         \multirow{5}{*}{Multi-Instance Retrieval}& \multirow{5}{*}{EPIC-KITCHENS-100} & \textit{Given a query video segment, how can captions be ranked so that those with a higher rank are more semantically relevant to the action in the query video segment?}\\
         & & (A) cut mozzarella into salad bowl. (B) cut mozzarella bag.\\
         & & (C) grab mozzarella. (D) put down mozzarella. (E) take mozzarella.\\
         \hline
         \multirow{4}{*}{Egocentric Navigation}& \multirow{4}{*}{VLN-QA} & \textit{This is a navigation video of an agent following instruction: ``Go left through the door.'' What is the next action it should take?}\\
         & & (A) turn left and move forward. (B) move forward. (C) stop.\\
         & & (D) turn right and move forward.\\
         \hline
         \multirow{5}{*}{Planning}& \multirow{5}{*}{EgoPlan} & \textit{Considering the progress shown in the video and my current observation in the last frame, what action should I take next for moving dough to baking tray?}\\
         & & (A) lay down dough on greaseproof paper. (B) put down chopstick.\\
         & & (C) put greaseproof paper on baking tray. (D) pick up greaseproof paper.\\
         \hline
         \multirow{5}{*}{Alignment}& \multirow{5}{*}{EgoMCQ} & \textit{Which description accurately reflects the human behavior in the video?}\\
         & & (A) I carry a yellow scoop of sugar from the sink with my gloved right hand. (B) I drop the plate in her right hand inside the cupboard. (C) I stop grating the coconut with a grater stool. (D) I holds lady w hair. (E) I drop my left hand by my side.\\
         \hline
    \end{tabular}
        \caption{Examples of tasks from different benchmarks.}

    \label{tab:sm_te}
\end{table*}

\subsubsection{EgoTaskQA}
EgoTaskQA~\cite{jia2022egotaskqa} is a large-scale egocentric video question-answering dataset designed to evaluate models' understanding of goal-oriented human tasks. It features 40K balanced question-answer pairs derived from 2K videos in the LEMMA dataset~\cite{jia2020lemma}, focusing on aspects such as action effects, intent, multi-agent collaboration, and object interactions. The dataset emphasizes reasoning types including spatial-temporal understanding, causal dependencies, and task planning, supported by 30K annotated state transitions. It includes a variety of question formats, such as binary and open-ended queries, to ensure a balanced and unbiased evaluation. EgoTaskQA is uniquely structured into two distinct subsets: \textbf{``direct" subset} contains randomly sampled questions, and \textbf{``indirect" subset} requires multi-step reasoning to derive the correct answer.

Due to the subtlety and ambiguity of many questions, open-domain QA in a zero-shot setting is difficult to accurately evaluate. Consequently, we structure it as a multiple-choice QA task. Specifically, we retain the original question from the testing set and select the four most similar incorrect answers from the answer set based on SentenceBERT~\cite{reimers2019sentence} similarity to the ground truth. These four, along with the correct answer, form a set of five candidates.

\subsubsection{Charades-Ego}
Charades-Ego~\cite{sigurdsson2018charades} is a large-scale dataset comprising 7,860 videos that capture everyday indoor activities from both first- and third-person perspectives. It contains 68.8 hours of video with 68,536 temporal annotations across 157 action classes. Derived from the original Charades dataset, Charades-Ego bridges first- and third-person video understanding by offering paired videos recorded simultaneously from both perspectives. Featuring 112 actors performing a diverse range of actions,  this dataset supports tasks such as video classification, localization, captioning, and cross-modal learning, making it a valuable resource for studying human activity from different viewpoints.

For model evaluation, we find that a multiple-choice QA format is more reliable and stable compared to direct answer generation. Accordingly, we create a comprehensive set of all actions and select the four most similar incorrect actions to each action label using SentenceBERT~\cite{reimers2019sentence} similarity. 
These four actions, combined with the correct answer, form a set of five candidate options for each sample. This approach is more effective than generating candidates with the LLMs, as it ensures that the options are realistic and require visual details for accurate differentiation. 

\begin{table*}[t]
    \renewcommand{\arraystretch}{1.1}
    \centering
    \begin{tabular}{cccccc}
        \hline
        \textbf{Benchmark} & \textbf{Metric} &\textbf{Initial} & \textbf{Stage 1}& \textbf{Stage 2}& \textbf{Stage 3}\\
         \hline
         \textbf{EgoSchema}&Acc.&\multicolumn{2}{c}{49.2}&56.7&\textbf{61.3}\\
         \rowcolor{gray!10} &Acc. (Closed)&\multicolumn{2}{c}{53.1}&54.4&\textbf{62.1}\\
         \rowcolor{gray!10} \multirow{-2}{*}{\textbf{QAEgo4D}}&Acc./Score (Open)&\multicolumn{2}{c}{21.8/1.9}&26.6/2.4&\textbf{28.3/2.7}\\
         \multirow{2}{*}{\textbf{EgoTaskQA}}&Acc. (Direct)&\multicolumn{2}{c}{35.4}&42.7&\textbf{44.7}\\
         &Acc. (Indirect)&\multicolumn{2}{c}{42.1}&47.8&\textbf{50.3}\\
         \rowcolor{gray!10} \textbf{Charades-Ego}& mAP&\multicolumn{2}{c}{62.3}&64.7&\textbf{70.9}\\
         \multirow{2}{*}{\textbf{EK-100 MIR}}&mAP&\multicolumn{2}{c}{41.1}&44.8&\textbf{49.7}\\
         &nDCG&\multicolumn{2}{c}{57.1}&61.6&\textbf{63.6}\\
         \rowcolor{gray!10} \textbf{EgoPlan Val}&Acc.&\multicolumn{2}{c}{35.1}&39.5&\textbf{42.7}\\
         \textbf{VLN-QA} &Acc.&\multicolumn{2}{c}{37.7}&41.0&\textbf{44.5}\\
         \rowcolor{gray!10}&Acc. (Inter)&\multicolumn{2}{c}{76.3}&82.7&\textbf{88.4}\\
         \rowcolor{gray!10} \multirow{-2}{*}{\textbf{EgoMCQ}}&Acc. (Intra)&\multicolumn{2}{c}{33.4}&37.7&\textbf{41.2}\\
         \hline
    \end{tabular}
       \caption{Performance comparison of Exo2Ego across different training stages on the multiple benchmarks. The QAEgo4D dataset is categorized into ``Closed" and ``Open" question answering, while the EgoTaskQA dataset includes ``Direct" and ``Indirect" subsets. In EgoMCQ, ``Inter" refers to cases where the candidates and input clips come from different videos, while ``Intra" indicates that they are from the same video, presenting a more challenging scenario. The best results are highlighted in bold.}
    \label{tab:sm_ds}
\end{table*}

\subsubsection{EPIC-KITCHENS-100}
EPIC-KITCHENS-100~\cite{damen2022rescaling} is a first-person vision dataset featuring over 100 hours of unscripted kitchen activities recorded in 45 kitchens across 4 cities. It contains 20 million frames, 90,000 action segments, and 20,000 unique activity descriptions, covering 97 verb classes and 300 noun classes. Captured using head-mounted cameras, the dataset supports a wide range of tasks, including action recognition, video classification, and multimodal learning, and is notable for its innovative ``Pause-and-Talk" annotation system. This dataset addresses five key challenges in advancing egocentric video understanding.

Since the data format of EPIC-KITCHENS-100 is similar to Charades-Ego, we follow the same processing steps as outlined for Charades-Ego.

\subsubsection{VLN-QA}
VLN-QA~\cite{krantz2020beyond} is a multiple-choice question-answering dataset designed to evaluate the navigation capabilities of vision-language systems in complex indoor environments. Built on the VLN-CE source, it contains thousands of curated question-answer pairs based on first-person video clips that simulate real-world navigation tasks. Each video lasts several minutes and challenges models to make decisions based on multimodal understanding, including visual perception, spatial reasoning, and path planning, all within intricate indoor settings.

This dataset is built by VideoChat2~\cite{li2024mvbench} using LLMs based on VLN-CE. Since it already fulfills the QA format requirements, we directly adopt the processed data provided by VideoChat2.

\subsubsection{EgoPlan}
EgoPlan~\cite{chen2023egoplan} is a comprehensive egocentric video benchmark designed to assess human-level planning capabilities in vision and language systems. Built from EPIC-KITCHENS~\cite{damen2022rescaling} and Ego4D~\cite{grauman2022ego4d}, it features 4,939 verified multiple-choice questions, covering 3,296 task goals and 3,185 action plans across 419 real-world scenes. The dataset involves interactions with 558 objects and 234 verbs, challenging models to select the correct next action plan based on visual observations from egocentric video clips. This makes EgoPlan a valuable tool for advancing multimodal models in real-world planning  scenarios.

The dataset format aligns with our task requirements, and we use the provided validation set for planning evaluation, as the true labels of the testing set are unavailable.

\subsubsection{EgoMCQ}
EgoMCQ~\cite{lin2022egocentric} is a multiple-choice question-answering dataset designed to assess video-text alignment in egocentric vision systems. Derived from Ego4D~\cite{grauman2022ego4d}, it includes 39,000 questions based on 468 hours of egocentric video covering a wide range of human activities. Each question involves selecting the correct video clip from five options based on a narration, with two settings: \textbf{``inter-video"}, for distinguishing between different videos, and \textbf{``intra-video"}, for fine-grained context within the same video. 

This dataset provides complete multiple-choice QA data. We use the video in each query as the visual input, and the original text options as candidates, to construct a video-text retrieval task. Since most methods can achieve 95\% accuracy (inter-video) under the original settings (1: 5, a query corresponds to 5 candidates), we increased the number of candidates to 10 (i.e., 1: 10) by adding similar text based on SentenceBERT~\cite{reimers2019sentence} similarity 
to highlight the differences among them.

\subsection{Example Presentation}\label{cep}
In Table~\ref{tab:sm_te}, we visually present an example from each benchmark. Each data source emphasizes a different aspect of task evaluation. For instance, QAEgo4D focuses on episodic memory, with questions like ``Where did I put the chopsticks?" that require precise recall of past details and comprehension to select the correct answer. On the other hand, the Egoschema source emphasizes the evaluation of the model's reasoning abilities, requiring a comprehensive understanding of the entire event, including its causes and consequences.

\section{More Experiments}\label{me}
\subsection{Hyperparameter Settings}\label{sm_hss}
Table~\ref{tab:sm_th} summarizes the hyperparameters used in different training stages. At each stage, we used 16-frame videos with a resolution of 336 $\times$ 336 as the visual input. 
The overall framework employs the AdamW~\cite{kingma2014adam} optimizer with momentum parameters $\beta_1=0.9$ and $\beta_2=0.999$, along with a weight decay of 0.02 to mitigate overfitting. The learning rate follows a cosine decay schedule for gradual reduction. In line with VideoLLaMA2, we enabled BF16 and TF32 precision while disabling FP16 precision to optimize performance using mixed-precision calculations on compatible hardware. To further enhance training efficiency, we implemented flash attention~\cite{dao2022flashattention}.

\subsection{Comparison Results}\label{dcr}


\subsubsection{Effect of stages}
To explore the impact of different stages on the overall performance of the model, we evaluated each stage separately and the results are shown in Table~\ref{tab:sm_ds}. From the results, we can observe that pre-training in the initialization provides basic egocentric video comprehension, while mapping learning in the Stage 2 effectively aligns data from different domains, leading to competitive results on all benchmarks. After Stage 3, the model's instruction-following capability is enhanced, enabling it to better understand instructions for downstream tasks, which significantly improves its performance. These results fully validate the effect and effectiveness of each stage in the overall pipeline.

\subsubsection{Effect of prompts}
Variations in prompt design can significantly influence the performance of MLLMs, as summarized in Table~\ref{tab:ab_pd}.  
The results in the second row indicate that a comprehensive system prompt, which explicitly emphasizes task requirements, enhances the effectiveness of task completion. Moreover, the third row shows that incorporating elements related to first-person video comprehension further improves the understanding of egocentric videos.
\begin{table}[ht]
    \renewcommand{\arraystretch}{1.0}
    \begin{tabular}{m{6.7cm}|m{0.7cm}<{\centering}}
        \hline
        \textbf{Prompt}& \textbf{Avg} \\
         \hline
         Carefully observe the video and choose the best option for the question.&54.5\\
         \hline
         Carefully watch the video and \textcolor{OliveGreen}{pay attention to the cause and sequence of events, the detail and movement of objects, and the action and pose of persons.} Select the best option that accurately addresses the question. & 55.2~\textcolor{OliveGreen}{$\uparrow$}\\
         \hline
         \textcolor{OliveGreen}{From a first-person perspective}, pay attention to the cause and sequence of events, \textcolor{OliveGreen}{the hands' movements and details with objects}, and the action and pose of persons. Select the best option that accurately addresses the question. &\textbf{55.6}~\textcolor{OliveGreen}{$\uparrow$} \\      
         \hline
    \end{tabular}
        \caption{Impact of different prompts on model performance. ``Avg'' represents the average results across all metrics in Figure 1 in the manuscript.}

    \label{tab:ab_pd}
\end{table}

\begin{figure}[ht]
  \centering
   \includegraphics[width=\linewidth]{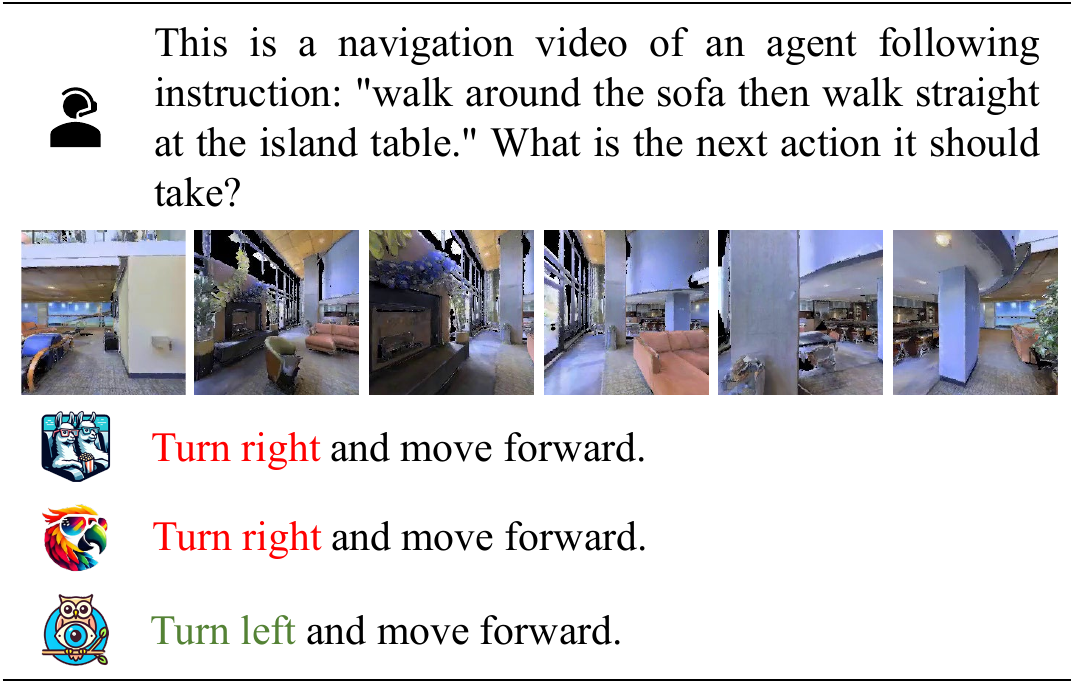}
   \caption{A qualitative example from the VLN-QA source.}
   \label{fig:supvisi}
\end{figure}

\begin{figure*}[!t]
    \centering
    \begin{subfigure}{\textwidth}
        \includegraphics[width=\linewidth]{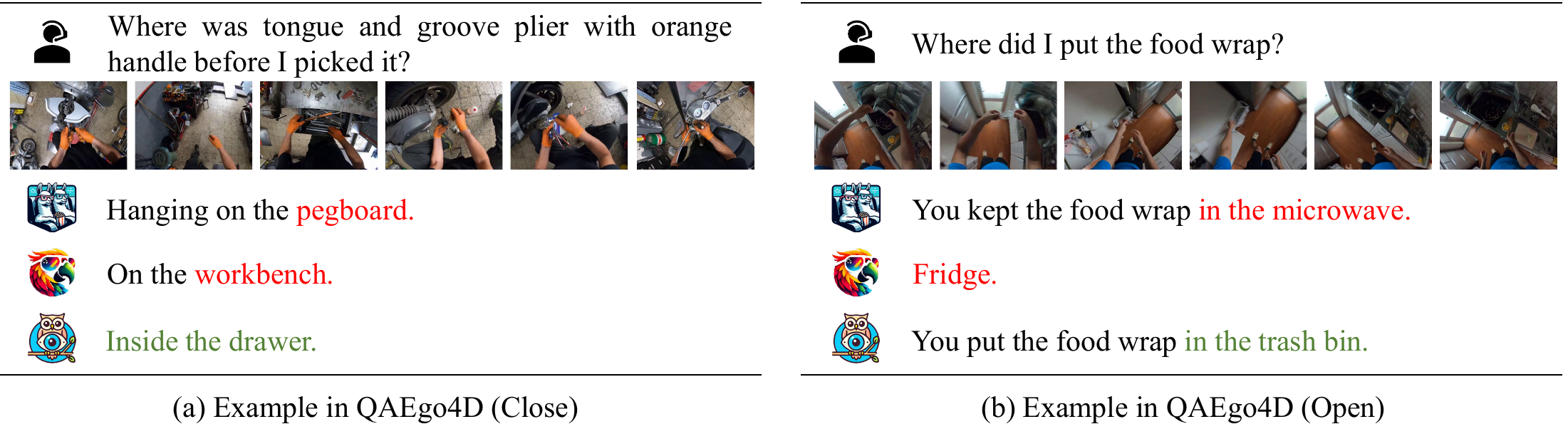} 
        \label{fig:subfig1}
    \end{subfigure}
    \hspace{0.05\textwidth}
    \begin{subfigure}{\textwidth}
        \includegraphics[width=\linewidth]{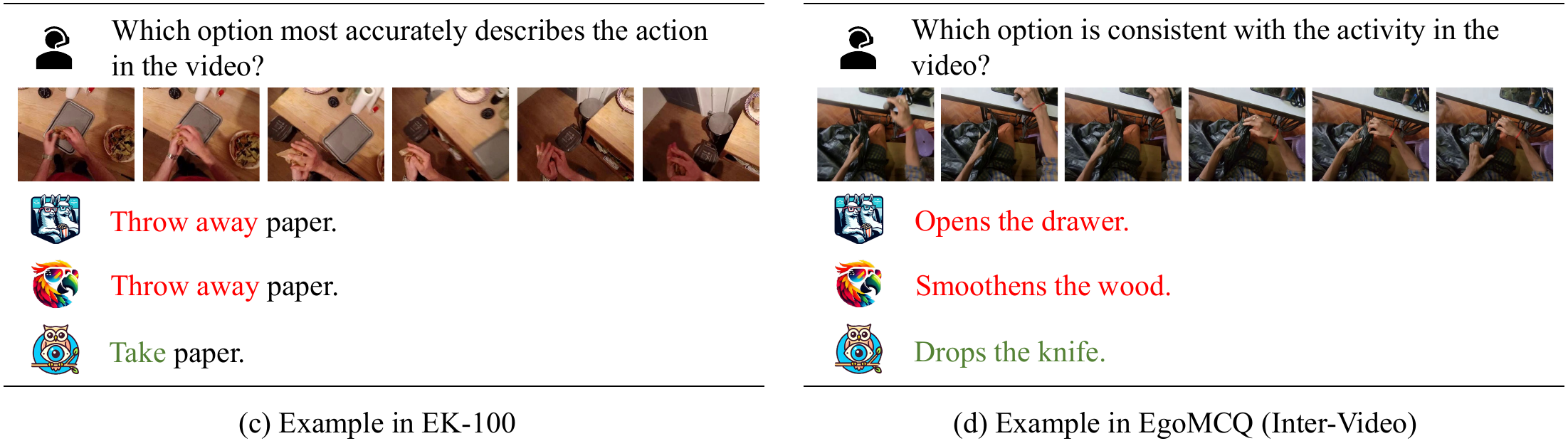} 
        \label{fig:subfig1}
    \end{subfigure}
    \hspace{0.05\textwidth}
    \begin{subfigure}{\textwidth}
        \includegraphics[width=\linewidth]{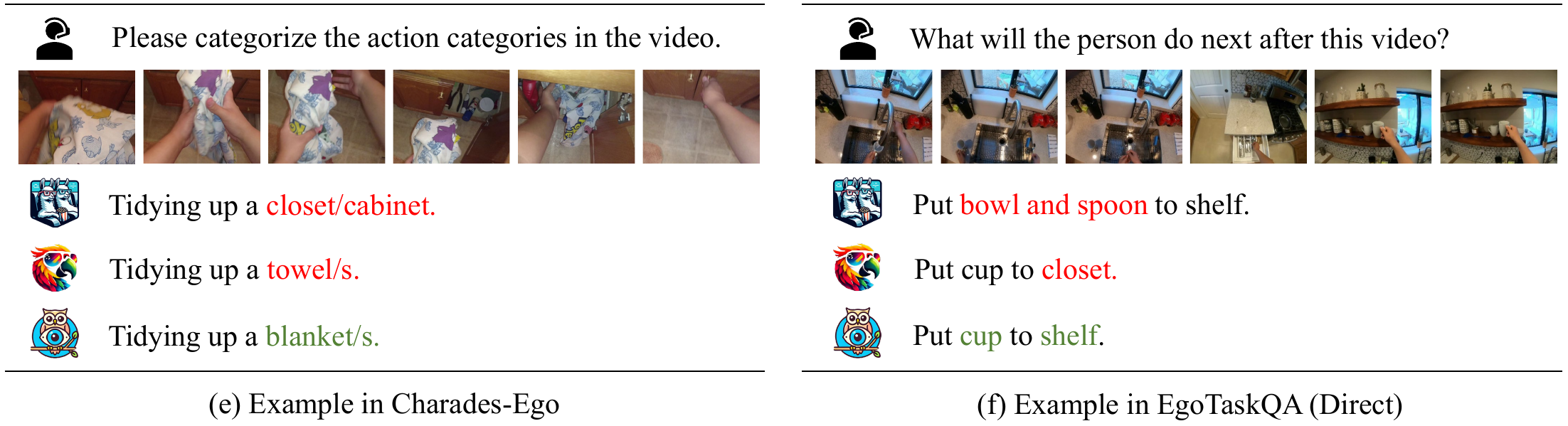} 
        \label{fig:subfig1}
    \end{subfigure}
    \hspace{0.05\textwidth}
    \begin{subfigure}{\textwidth}
        \includegraphics[width=\linewidth]{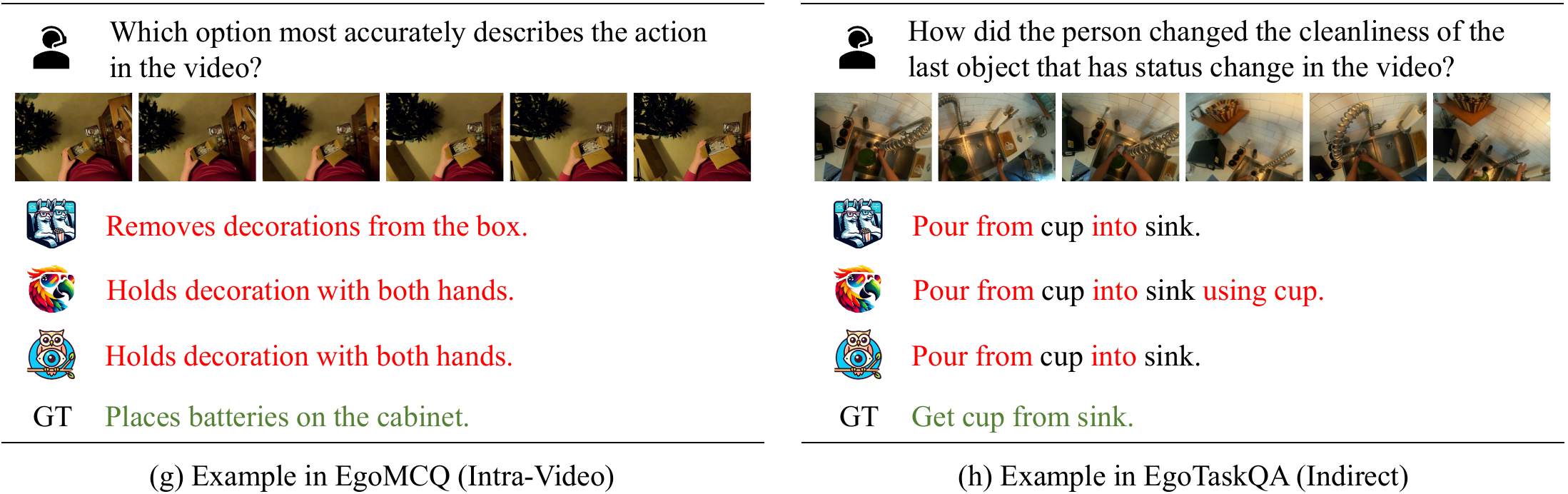} 
        \label{fig:subfig1}
    \end{subfigure}
    
    
    \caption{Additional qualitative examples from all benchmarks. ``GT" denotes the ground truth.}
    \label{fig:supvisah}
\end{figure*}

\begin{table}[t]
    \renewcommand{\arraystretch}{1.0}
    \centering
    \begin{tabular}{cccc}
        \hline
        \textbf{Config} &\textbf{Initial}& \textbf{Stage 1\&2}&\textbf{Stage 3}\\
         \hline
         Global Batch size & 512 & 256 & 64\\
         Local Batch size & 8 & 8 & 8\\
         Learning rate & 1e-3&1e-4&2e-5\\
         Warmup ratio & 0.1&0.03&0.03\\
         Total epoch & 5 & 2 & 1\\
         \hline
    \end{tabular}
       \caption{Hyperparameter settings for different training stages.}

    \label{tab:sm_th}
\end{table}

\subsection{Qualitative Analysis}\label{mqa}
We present additional qualitative examples from various data sources, as shown in Figures~\ref{fig:supvisi} and~\ref{fig:supvisah}. A detailed analysis is summarized below:
\begin{itemize}
    \item In Figure~\ref{fig:supvisi}, both VideoChat2 and VideoLLaMA2 provide the incorrect navigation instruction,  ``turn right and move forward", while our Exo2Ego correctly suggests ``turn left and move forward" to achieve the given goal.
    \item Figure~\ref{fig:supvisah}(a) shows the results of closed-ended QA on the QAEgo4D dataset. Compared to the two optimal baselines, our approach accurately pinpoints the spatiotemporal location of the object ``tongue and groove plier with orange handle" and selects the correct answer. In the case of open-ended QA presented in Figure~\ref{fig:supvisah}(b), our Exo2Ego also identifies the action of ``food wrap being thrown into the trash bin" and generates the correct response. 
    \item In Figure~\ref{fig:supvisah}(c), which is from the EPIC-KITCHENS-100 dataset, the two baselines fail to interpret the visual details and instead rely on common sense, assuming that standing by a trash bin while holding paper implies discarding it. In contrast, our method focuses on and correctly interprets the finer details of hand-object interactions. In Figure~\ref{fig:supvisah}(d), our model accurately understands the current activity, whereas existing methods fail to focus on hand movements and instead make judgments based solely on the objects present in the scene.

    \item Figure~\ref{fig:supvisah} also presents an example of action recognition from Charades-Ego. In the first instance, all three methods correctly identify ``Tidying up", but only our method understands which object is being manipulated. In the second example on the same row, the answers from all three methods are relatively similar, but considering the understanding of finer details, the selection by Exo2Ego is the most appropriate.
    
    \item We also provide two failure cases at the end of Figure~\ref{fig:supvisah}. In Figure~\ref{fig:supvisah}(g), both our Exo2Ego model and the two baseline methods select the incorrect answer, likely because the object ``battery" is too small in the current perspective to be accurately identified. In Figure~\ref{fig:supvisah}(h), all the answers are incorrect, likely due to significant ambiguity in the \textit{indirect} setting from the EgoTaskQA dataset, making it challenging to interpret the scenario correctly. 
\end{itemize}

\section{Limitations}\label{limit}
The current scale of egocentric data used for training and evaluation is relatively small, with limited diversity, which may lead to insufficient learning and incomplete validation. Additionally, employing more advanced teacher models (e.g., Qwen2-VL~\cite{wang2024qwen2}) and investigating more effective mapping techniques could facilitate the transfer of more precise exocentric knowledge, thereby enhancing the understanding of egocentric videos.


%% file: main.bib
@String(CVPR= {IEEE Conf. Comput. Vis. Pattern Recog.})

@String(CVPR  = {CVPR})

@inproceedings{pramanick2023egovlpv2,
  title={Egovlpv2: Egocentric video-language pre-training with fusion in the backbone},
  author={Pramanick, Shraman and Song, Yale and Nag, Sayan and Lin, Kevin Qinghong and Shah, Hardik and Shou, Mike Zheng and Chellappa, Rama and Zhang, Pengchuan},
  booktitle={Proceedings of the IEEE/CVF International Conference on Computer Vision},
  pages={5285--5297},
  year={2023}
}

@article{lin2022egocentric,
  title={Egocentric video-language pretraining},
  author={Lin, Kevin Qinghong and Wang, Jinpeng and Soldan, Mattia and Wray, Michael and Yan, Rui and Xu, Eric Z and Gao, Difei and Tu, Rong-Cheng and Zhao, Wenzhe and Kong, Weijie and others},
  journal={Advances in Neural Information Processing Systems},
  volume={35},
  pages={7575--7586},
  year={2022}
}

@inproceedings{di2024grounded,
  title={Grounded Question-Answering in Long Egocentric Videos},
  author={Di, Shangzhe and Xie, Weidi},
  booktitle={Proceedings of the IEEE/CVF Conference on Computer Vision and Pattern Recognition},
  pages={12934--12943},
  year={2024}
}

@article{li2023llama,
  title={Llama-vid: An image is worth 2 tokens in large language models},
  author={Li, Yanwei and Wang, Chengyao and Jia, Jiaya},
  journal={arXiv preprint arXiv:2311.17043},
  year={2023}
}

@article{cheng2024videollama,
  title={VideoLLaMA 2: Advancing Spatial-Temporal Modeling and Audio Understanding in Video-LLMs},
  author={Cheng, Zesen and Leng, Sicong and Zhang, Hang and Xin, Yifei and Li, Xin and Chen, Guanzheng and Zhu, Yongxin and Zhang, Wenqi and Luo, Ziyang and Zhao, Deli and others},
  journal={arXiv preprint arXiv:2406.07476},
  year={2024}
}

@article{lin2023video,
  title={Video-llava: Learning united visual representation by alignment before projection},
  author={Lin, Bin and Zhu, Bin and Ye, Yang and Ning, Munan and Jin, Peng and Yuan, Li},
  journal={arXiv preprint arXiv:2311.10122},
  year={2023}
}

@inproceedings{li2024mvbench,
  title={Mvbench: A comprehensive multi-modal video understanding benchmark},
  author={Li, Kunchang and Wang, Yali and He, Yinan and Li, Yizhuo and Wang, Yi and Liu, Yi and Wang, Zun and Xu, Jilan and Chen, Guo and Luo, Ping and others},
  booktitle={Proceedings of the IEEE/CVF Conference on Computer Vision and Pattern Recognition},
  pages={22195--22206},
  year={2024}
}

@article{maaz2023video,
  title={Video-chatgpt: Towards detailed video understanding via large vision and language models},
  author={Maaz, Muhammad and Rasheed, Hanoona and Khan, Salman and Khan, Fahad Shahbaz},
  journal={arXiv preprint arXiv:2306.05424},
  year={2023}
}

@article{touvron2023llama,
  title={Llama 2: Open foundation and fine-tuned chat models},
  author={Touvron, Hugo and Martin, Louis and Stone, Kevin and Albert, Peter and Almahairi, Amjad and Babaei, Yasmine and Bashlykov, Nikolay and Batra, Soumya and Bhargava, Prajjwal and Bhosale, Shruti and others},
  journal={arXiv preprint arXiv:2307.09288},
  year={2023}
}

@article{alayrac2022flamingo,
  title={Flamingo: a visual language model for few-shot learning},
  author={Alayrac, Jean-Baptiste and Donahue, Jeff and Luc, Pauline and Miech, Antoine and Barr, Iain and Hasson, Yana and Lenc, Karel and Mensch, Arthur and Millican, Katherine and Reynolds, Malcolm and others},
  journal={Advances in neural information processing systems},
  volume={35},
  pages={23716--23736},
  year={2022}
}

@inproceedings{li2023blip,
  title={Blip-2: Bootstrapping language-image pre-training with frozen image encoders and large language models},
  author={Li, Junnan and Li, Dongxu and Savarese, Silvio and Hoi, Steven},
  booktitle={International conference on machine learning},
  pages={19730--19742},
  year={2023},
  organization={PMLR}
}

@article{liu2024visual,
  title={Visual instruction tuning},
  author={Liu, Haotian and Li, Chunyuan and Wu, Qingyang and Lee, Yong Jae},
  journal={Advances in neural information processing systems},
  volume={36},
  year={2024}
}

@inproceedings{NEURIPS2023_9a6a435e,
 author = {Dai, Wenliang and Li, Junnan and LI, DONGXU and Tiong, Anthony and Zhao, Junqi and Wang, Weisheng and Li, Boyang and Fung, Pascale N and Hoi, Steven},
 booktitle = {Advances in Neural Information Processing Systems},
 editor = {A. Oh and T. Naumann and A. Globerson and K. Saenko and M. Hardt and S. Levine},
 pages = {49250--49267},
 publisher = {Curran Associates, Inc.},
 title = {InstructBLIP: Towards General-purpose Vision-Language Models with Instruction Tuning},
 url = {https://proceedings.neurips.cc/paper_files/paper/2023/file/9a6a435e75419a836fe47ab6793623e6-Paper-Conference.pdf},
 volume = {36},
 year = {2023}
}

@article{damen2020epic,
  title={The epic-kitchens dataset: Collection, challenges and baselines},
  author={Damen, Dima and Doughty, Hazel and Farinella, Giovanni Maria and Fidler, Sanja and Furnari, Antonino and Kazakos, Evangelos and Moltisanti, Davide and Munro, Jonathan and Perrett, Toby and Price, Will and others},
  journal={IEEE Transactions on Pattern Analysis and Machine Intelligence},
  volume={43},
  number={11},
  pages={4125--4141},
  year={2020},
  publisher={IEEE}
}

@article{sigurdsson2018charades,
  title={Charades-ego: A large-scale dataset of paired third and first person videos},
  author={Sigurdsson, Gunnar A and Gupta, Abhinav and Schmid, Cordelia and Farhadi, Ali and Alahari, Karteek},
  journal={arXiv preprint arXiv:1804.09626},
  year={2018}
}

@inproceedings{grauman2022ego4d,
  title={Ego4d: Around the world in 3,000 hours of egocentric video},
  author={Grauman, Kristen and Westbury, Andrew and Byrne, Eugene and Chavis, Zachary and Furnari, Antonino and Girdhar, Rohit and Hamburger, Jackson and Jiang, Hao and Liu, Miao and Liu, Xingyu and others},
  booktitle={Proceedings of the IEEE/CVF Conference on Computer Vision and Pattern Recognition},
  pages={18995--19012},
  year={2022}
}

@inproceedings{jiang2023full,
  title={Full-body articulated human-object interaction},
  author={Jiang, Nan and Liu, Tengyu and Cao, Zhexuan and Cui, Jieming and Zhang, Zhiyuan and Chen, Yixin and Wang, He and Zhu, Yixin and Huang, Siyuan},
  booktitle={Proceedings of the IEEE/CVF International Conference on Computer Vision},
  pages={9365--9376},
  year={2023}
}

@inproceedings{wang2023ego,
  title={Ego-only: Egocentric action detection without exocentric transferring},
  author={Wang, Huiyu and Singh, Mitesh Kumar and Torresani, Lorenzo},
  booktitle={Proceedings of the IEEE/CVF International Conference on Computer Vision},
  pages={5250--5261},
  year={2023}
}

@inproceedings{shiota2024egocentric,
  title={Egocentric action recognition by capturing hand-object contact and object state},
  author={Shiota, Tsukasa and Takagi, Motohiro and Kumagai, Kaori and Seshimo, Hitoshi and Aono, Yushi},
  booktitle={Proceedings of the IEEE/CVF Winter Conference on Applications of Computer Vision},
  pages={6541--6551},
  year={2024}
}

@inproceedings{ragusa2023stillfast,
  title={Stillfast: An end-to-end approach for short-term object interaction anticipation},
  author={Ragusa, Francesco and Farinella, Giovanni Maria and Furnari, Antonino},
  booktitle={Proceedings of the IEEE/CVF Conference on Computer Vision and Pattern Recognition},
  pages={3636--3645},
  year={2023}
}

@inproceedings{ramakrishnan2023naq,
  title={Naq: Leveraging narrations as queries to supervise episodic memory},
  author={Ramakrishnan, Santhosh Kumar and Al-Halah, Ziad and Grauman, Kristen},
  booktitle={Proceedings of the IEEE/CVF Conference on Computer Vision and Pattern Recognition},
  pages={6694--6703},
  year={2023}
}

@inproceedings{akiva2023self,
  title={Self-supervised object detection from egocentric videos},
  author={Akiva, Peri and Huang, Jing and Liang, Kevin J and Kovvuri, Rama and Chen, Xingyu and Feiszli, Matt and Dana, Kristin and Hassner, Tal},
  booktitle={Proceedings of the IEEE/CVF International Conference on Computer Vision},
  pages={5225--5237},
  year={2023}
}

@article{jia2022egotaskqa,
  title={Egotaskqa: Understanding human tasks in egocentric videos},
  author={Jia, Baoxiong and Lei, Ting and Zhu, Song-Chun and Huang, Siyuan},
  journal={Advances in Neural Information Processing Systems},
  volume={35},
  pages={3343--3360},
  year={2022}
}

@article{zhang2024hcqa,
  title={HCQA@ Ego4D EgoSchema Challenge 2024},
  author={Zhang, Haoyu and Xie, Yuquan and Feng, Yisen and Li, Zaijing and Liu, Meng and Nie, Liqiang},
  journal={arXiv preprint arXiv:2406.15771},
  year={2024}
}

@article{feng2024objectnlq,
  title={ObjectNLQ@ Ego4D Episodic Memory Challenge 2024},
  author={Feng, Yisen and Zhang, Haoyu and Xie, Yuquan and Li, Zaijing and Liu, Meng and Nie, Liqiang},
  journal={arXiv preprint arXiv:2406.15778},
  year={2024}
}

@inproceedings{wang2023learning,
  title={Learning from semantic alignment between unpaired multiviews for egocentric video recognition},
  author={Wang, Qitong and Zhao, Long and Yuan, Liangzhe and Liu, Ting and Peng, Xi},
  booktitle={Proceedings of the IEEE/CVF International Conference on Computer Vision},
  pages={3307--3317},
  year={2023}
}

@article{xue2023learning,
  title={Learning fine-grained view-invariant representations from unpaired ego-exo videos via temporal alignment},
  author={Xue, Zihui Sherry and Grauman, Kristen},
  journal={Advances in Neural Information Processing Systems},
  volume={36},
  pages={53688--53710},
  year={2023}
}

@inproceedings{li2021ego,
  title={Ego-exo: Transferring visual representations from third-person to first-person videos},
  author={Li, Yanghao and Nagarajan, Tushar and Xiong, Bo and Grauman, Kristen},
  booktitle={Proceedings of the IEEE/CVF Conference on Computer Vision and Pattern Recognition},
  pages={6943--6953},
  year={2021}
}

@article{li2021eye,
  title={In the eye of the beholder: Gaze and actions in first person video},
  author={Li, Yin and Liu, Miao and Rehg, James M},
  journal={IEEE transactions on pattern analysis and machine intelligence},
  volume={45},
  number={6},
  pages={6731--6747},
  year={2021},
  publisher={IEEE}
}

@inproceedings{goyal2017something,
  title={The" something something" video database for learning and evaluating visual common sense},
  author={Goyal, Raghav and Ebrahimi Kahou, Samira and Michalski, Vincent and Materzynska, Joanna and Westphal, Susanne and Kim, Heuna and Haenel, Valentin and Fruend, Ingo and Yianilos, Peter and Mueller-Freitag, Moritz and others},
  booktitle={Proceedings of the IEEE international conference on computer vision},
  pages={5842--5850},
  year={2017}
}

@InProceedings{Majumdar_2024_CVPR,
    author    = {Majumdar, Arjun and Ajay, Anurag and Zhang, Xiaohan and Putta, Pranav and Yenamandra, Sriram and Henaff, Mikael and Silwal, Sneha and Mcvay, Paul and Maksymets, Oleksandr and Arnaud, Sergio and Yadav, Karmesh and Li, Qiyang and Newman, Ben and Sharma, Mohit and Berges, Vincent and Zhang, Shiqi and Agrawal, Pulkit and Bisk, Yonatan and Batra, Dhruv and Kalakrishnan, Mrinal and Meier, Franziska and Paxton, Chris and Sax, Alexander and Rajeswaran, Aravind},
    title     = {OpenEQA: Embodied Question Answering in the Era of Foundation Models},
    booktitle = {Proceedings of the IEEE/CVF Conference on Computer Vision and Pattern Recognition (CVPR)},
    month     = {June},
    year      = {2024},
    pages     = {16488-16498}
}

@inproceedings{huang2024egoexolearn,
  title={EgoExoLearn: A Dataset for Bridging Asynchronous Ego-and Exo-centric View of Procedural Activities in Real World},
  author={Huang, Yifei and Chen, Guo and Xu, Jilan and Zhang, Mingfang and Yang, Lijin and Pei, Baoqi and Zhang, Hongjie and Dong, Lu and Wang, Yali and Wang, Limin and others},
  booktitle={Proceedings of the IEEE/CVF Conference on Computer Vision and Pattern Recognition},
  pages={22072--22086},
  year={2024}
}

@ARTICLE{10.3389/fpsyg.2015.00875,
AUTHOR={Johnson, Mark },
TITLE={Embodied understanding},
JOURNAL={Frontiers in Psychology},
VOLUME={6},
YEAR={2015},
URL={https://www.frontiersin.org/journals/psychology/articles/10.3389/fpsyg.2015.00875},
DOI={10.3389/fpsyg.2015.00875},
ISSN={1664-1078}
}

@inproceedings{grauman2024ego,
  title={Ego-exo4d: Understanding skilled human activity from first-and third-person perspectives},
  author={Grauman, Kristen and Westbury, Andrew and Torresani, Lorenzo and Kitani, Kris and Malik, Jitendra and Afouras, Triantafyllos and Ashutosh, Kumar and Baiyya, Vijay and Bansal, Siddhant and Boote, Bikram and others},
  booktitle={Proceedings of the IEEE/CVF Conference on Computer Vision and Pattern Recognition},
  pages={19383--19400},
  year={2024}
}

@article{dao2022flashattention,
  title={Flashattention: Fast and memory-efficient exact attention with io-awareness},
  author={Dao, Tri and Fu, Dan and Ermon, Stefano and Rudra, Atri and R{\'e}, Christopher},
  journal={Advances in Neural Information Processing Systems},
  volume={35},
  pages={16344--16359},
  year={2022}
}

@article{mangalam2023egoschema,
  title={Egoschema: A diagnostic benchmark for very long-form video language understanding},
  author={Mangalam, Karttikeya and Akshulakov, Raiymbek and Malik, Jitendra},
  journal={Advances in Neural Information Processing Systems},
  volume={36},
  pages={46212--46244},
  year={2023}
}

@inproceedings{barmann2022did,
  title={Where did i leave my keys?-episodic-memory-based question answering on egocentric videos},
  author={B{\"a}rmann, Leonard and Waibel, Alex},
  booktitle={Proceedings of the IEEE/CVF Conference on Computer Vision and Pattern Recognition},
  pages={1560--1568},
  year={2022}
}

@article{damen2022rescaling,
  title={Rescaling egocentric vision: Collection, pipeline and challenges for epic-kitchens-100},
  author={Damen, Dima and Doughty, Hazel and Farinella, Giovanni Maria and Furnari, Antonino and Kazakos, Evangelos and Ma, Jian and Moltisanti, Davide and Munro, Jonathan and Perrett, Toby and Price, Will and others},
  journal={International Journal of Computer Vision},
  pages={1--23},
  year={2022},
  publisher={Springer}
}

@inproceedings{krantz2020beyond,
  title={Beyond the nav-graph: Vision-and-language navigation in continuous environments},
  author={Krantz, Jacob and Wijmans, Erik and Majumdar, Arjun and Batra, Dhruv and Lee, Stefan},
  booktitle={Computer Vision--ECCV 2020: 16th European Conference, Glasgow, UK, August 23--28, 2020, Proceedings, Part XXVIII 16},
  pages={104--120},
  year={2020},
  organization={Springer}
}

@article{chen2023egoplan,
  title={EgoPlan-Bench: Benchmarking Egocentric Embodied Planning with Multimodal Large Language Models},
  author={Chen, Yi and Ge, Yuying and Ge, Yixiao and Ding, Mingyu and Li, Bohao and Wang, Rui and Xu, Ruifeng and Shan, Ying and Liu, Xihui},
  journal={arXiv preprint arXiv:2312.06722},
  year={2023}
}

@inproceedings{shen2024progress,
  title={Progress-aware online action segmentation for egocentric procedural task videos},
  author={Shen, Yuhan and Elhamifar, Ehsan},
  booktitle={Proceedings of the IEEE/CVF Conference on Computer Vision and Pattern Recognition},
  pages={18186--18197},
  year={2024}
}

@inproceedings{millerdurai2024eventego3d,
  title={Eventego3d: 3d human motion capture from egocentric event streams},
  author={Millerdurai, Christen and Akada, Hiroyasu and Wang, Jian and Luvizon, Diogo and Theobalt, Christian and Golyanik, Vladislav},
  booktitle={Proceedings of the IEEE/CVF Conference on Computer Vision and Pattern Recognition},
  pages={1186--1195},
  year={2024}
}

@inproceedings{luo2025put,
  title={Put myself in your shoes: Lifting the egocentric perspective from exocentric videos},
  author={Luo, Mi and Xue, Zihui and Dimakis, Alex and Grauman, Kristen},
  booktitle={European Conference on Computer Vision},
  pages={407--425},
  year={2025},
  organization={Springer}
}

@article{dou2024unlocking,
  title={Unlocking exocentric video-language data for egocentric video representation learning},
  author={Dou, Zi-Yi and Yang, Xitong and Nagarajan, Tushar and Wang, Huiyu and Huang, Jing and Peng, Nanyun and Kitani, Kris and Chu, Fu-Jen},
  journal={arXiv preprint arXiv:2408.03567},
  year={2024}
}

@article{li2024egoexo,
  title={EgoExo-Fitness: Towards Egocentric and Exocentric Full-Body Action Understanding},
  author={Li, Yuan-Ming and Huang, Wei-Jin and Wang, An-Lan and Zeng, Ling-An and Meng, Jing-Ke and Zheng, Wei-Shi},
  journal={arXiv preprint arXiv:2406.08877},
  year={2024}
}

@article{truong2024cross,
  title={Cross-view action recognition understanding from exocentric to egocentric perspective},
  author={Truong, Thanh-Dat and Luu, Khoa},
  journal={Neurocomputing},
  pages={128731},
  year={2024},
  publisher={Elsevier}
}

@article{shi2024cognition,
  title={Cognition Transferring and Decoupling for Text-supervised Egocentric Semantic Segmentation},
  author={Shi, Zhaofeng and Qiu, Heqian and Wang, Lanxiao and Meng, Fanman and Wu, Qingbo and Li, Hongliang},
  journal={arXiv preprint arXiv:2410.01341},
  year={2024}
}

@inproceedings{jia2020lemma,
  title={LEMMA: A Multi-view Dataset for LE arning M ulti-agent M ulti-task A ctivities},
  author={Jia, Baoxiong and Chen, Yixin and Huang, Siyuan and Zhu, Yixin and Zhu, Song-chun},
  booktitle={European Conference on Computer Vision},
  pages={767--786},
  year={2020},
  organization={Springer}
}

@article{reimers2019sentence,
  title={Sentence-BERT: Sentence Embeddings using Siamese BERT-Networks},
  author={Reimers, N},
  journal={arXiv preprint arXiv:1908.10084},
  year={2019}
}

@article{kingma2014adam,
  title={Adam: A method for stochastic optimization},
  author={Kingma, Diederik P},
  journal={arXiv preprint arXiv:1412.6980},
  year={2014}
}

@article{wang2024qwen2,
  title={Qwen2-vl: Enhancing vision-language model's perception of the world at any resolution},
  author={Wang, Peng and Bai, Shuai and Tan, Sinan and Wang, Shijie and Fan, Zhihao and Bai, Jinze and Chen, Keqin and Liu, Xuejing and Wang, Jialin and Ge, Wenbin and others},
  journal={arXiv preprint arXiv:2409.12191},
  year={2024}
}

@inproceedings{xu2025egocentric,
  title={Do Egocentric Video-Language Models Truly Understand Hand-Object Interactions?},
  author={Xu, Boshen and Wang, Ziheng and Du, Yang and Song, Zhinan and Zheng, Sipeng and Jin, Qin},
  booktitle={The Thirteenth International Conference on Learning Representations},
  year={2025}
}

@article{bai2025qwen2,
  title={Qwen2. 5-vl technical report},
  author={Bai, Shuai and Chen, Keqin and Liu, Xuejing and Wang, Jialin and Ge, Wenbin and Song, Sibo and Dang, Kai and Wang, Peng and Wang, Shijie and Tang, Jun and others},
  journal={arXiv preprint arXiv:2502.13923},
  year={2025}
}

@article{chu2025intention,
  title={Intention-Guided Cognitive Reasoning for Egocentric Long-Term Action Anticipation},
  author={Chu, Qiaohui and Zhang, Haoyu and Liu, Meng and Feng, Yisen and Shi, Haoxiang and Nie, Liqiang},
  journal={arXiv preprint arXiv:2508.01742},
  year={2025}
}

@article{chu2025technical,
  title={Technical report for ego4d long-term action anticipation challenge 2025},
  author={Chu, Qiaohui and Zhang, Haoyu and Feng, Yisen and Liu, Meng and Guan, Weili and Wang, Yaowei and Nie, Liqiang},
  journal={arXiv preprint arXiv:2506.02550},
  year={2025}
}

@article{wang2022siamese,
  title={Siamese alignment network for weakly supervised video moment retrieval},
  author={Wang, Yunxiao and Liu, Meng and Wei, Yinwei and Cheng, Zhiyong and Wang, Yinglong and Nie, Liqiang},
  journal={IEEE Transactions on Multimedia},
  volume={25},
  pages={3921--3933},
  year={2022},
  publisher={IEEE}
}

@article{wang2025time,
  title={Time: Temporal-sensitive multi-dimensional instruction tuning and benchmarking for video-llms},
  author={Wang, Yunxiao and Liu, Meng and Shao, Rui and Zhang, Haoyu and Wen, Bin and Yang, Fan and Gao, Tingting and Zhang, Di and Nie, Liqiang},
  journal={arXiv preprint arXiv:2503.09994},
  year={2025}
}

@inproceedings{zhang2021multimodal,
title={Multimodal dialog system: Relational graph-based context-aware question understanding},
author={Zhang, Haoyu and Liu, Meng and Gao, Zan and Lei, Xiaoqiang and Wang, Yinglong and Nie, Liqiang},
booktitle={Proceedings of the 29th ACM international conference on multimedia},
pages={695--703},
year={2021}
}

@InProceedings{pmlr-v235-zhang24aj,
title = 	 {Multi-Factor Adaptive Vision Selection for Egocentric Video Question Answering},
author =       {Zhang, Haoyu and Liu, Meng and Liu, Zixin and Song, Xuemeng and Wang, Yaowei and Nie, Liqiang},
booktitle = 	 {Proceedings of the 41st International Conference on Machine Learning},
pages = 	 {59310--59328},
year = 	 {2024},
volume = 	 {235},
publisher =    {PMLR}
}

@inproceedings{zhang2025spatial,
title={Spatial Understanding from Videos: Structured Prompts Meet Simulation Data},
author={Zhang, Haoyu and Liu, Meng and Li, Zaijing and Wen, Haokun and Guan, Weili and Wang, Yaowei and Nie, Liqiang},
booktitle = {Advances in Neural Information Processing Systems},
year      = {2025},
pages     = {1-16}
}

@article{zhang2023attribute,
title={Attribute-guided collaborative learning for partial person re-identification},
author={Zhang, Haoyu and Liu, Meng and Li, Yuhong and Yan, Ming and Gao, Zan and Chang, Xiaojun and Nie, Liqiang},
journal={IEEE Transactions on Pattern Analysis and Machine Intelligence},
volume={45},
number={12},
pages={14144--14160},
year={2023},
publisher={IEEE}
}

@InProceedings{Feng_2025_CVPR,
    author    = {Feng, Yisen and Zhang, Haoyu and Liu, Meng and Guan, Weili and Nie, Liqiang},
    title     = {Object-Shot Enhanced Grounding Network for Egocentric Video},
    booktitle = {Proceedings of the IEEE/CVF Conference on Computer Vision and Pattern Recognition (CVPR)},
    month     = {June},
    year      = {2025},
    pages     = {24190-24200}
}

@article{DBLP:journals/pami/WenSYWGN24,
  author       = {Haokun Wen and
                  Xuemeng Song and
                  Jianhua Yin and
                  Jianlong Wu and
                  Weili Guan and
                  Liqiang Nie},
  title        = {Self-Training Boosted Multi-Factor Matching Network for Composed Image
                  Retrieval},
  journal      = {{IEEE} Trans. Pattern Anal. Mach. Intell.},
  volume       = {46},
  number       = {5},
  pages        = {3665--3678},
  year         = {2024}
}
